\documentclass[11pt]{article}

\usepackage[final]{acl}

\usepackage{times}
\usepackage{array}
\usepackage{longtable}
\usepackage{latexsym}
\usepackage{amsmath}
\usepackage{graphicx}
\usepackage{multirow}
\usepackage{array}
\usepackage{graphicx}
\usepackage{microtype}
\usepackage{amsfonts}
\usepackage{mathrsfs}
\usepackage{multirow}
\usepackage{amssymb}
\usepackage{bbding}
\usepackage{subfigure} 
\usepackage{CJKutf8}
\usepackage{colortbl}
\usepackage{color}
\usepackage{booktabs}
\usepackage{makecell}
\usepackage{calligra}
\usepackage{algorithm} 
\usepackage{subcaption}
\usepackage{subfigure}
\usepackage[T1]{fontenc} 

\usepackage[utf8]{inputenc}

\usepackage{microtype}

\usepackage{inconsolata}

\usepackage{graphicx}

\title{CHisAgent: A Multi-Agent Framework for Event Taxonomy Construction in Ancient Chinese Cultural Systems}

\author{
\textbf{Xuemei Tang\textsuperscript{1}}\quad
\textbf{Chengxi Yan \textsuperscript{2}\thanks{Corresponding Author}}\quad
\textbf{Jinghang Gu\textsuperscript{1}\footnotemark[1]}\quad
\textbf{Chu-Ren Huang\textsuperscript{1}
}
\\
\textsuperscript{1}The Hong Kong Polytechnic University \\
\textsuperscript{2}Renmin University of China\\
\texttt{xuemeitang00@gmail.com, 20218113@ruc.edu.cn}\\
\texttt{\{jinghang.gu, churen.huang\}@polyu.edu.hk}
}

\begin{document}
\begin{CJK}{UTF8}{bsmi}

\maketitle
\begin{abstract}
Due to the long temporal span and highly diverse expressions in historical documents, large language models (LLMs) still struggle with historical and cultural reasoning in Chinese history. Although taxonomic structures can effectively organize historical knowledge and facilitate knowledge-rich applications, existing historical taxonomies remain limited due to the high cost of manual construction. To address this limitation, we propose \textbf{CHisAgent}, a multi-agent LLM framework for automated historical taxonomy induction in ancient Chinese history.
CHisAgent explicitly decomposes taxonomy construction into three role-specialized stages: a bottom-up \textit{Inducer} that derives an initial hierarchy from raw historical corpora, a top-down \textit{Expander} that introduces missing intermediate concepts using LLM world knowledge, and an evidence-guided \textit{Enricher} that integrates external structured historical resources to ensure faithfulness.
Extensive reference-free and reference-based evaluations show that the collaboration among these agents improves taxonomy coverage and hierarchical completeness, while further analysis suggests that the induced taxonomy facilitates cross-cultural alignment of historical concepts. Code and experimental data are available at ~\url{https://github.com/tangxuemei1995/CHisAgent}.
\end{abstract}

\section{Introduction}

Recently, LLMs have achieved strong performance in real-world understanding tasks such as emotion understanding~\cite{belay-etal-2025-culemo} and scene planning~\cite{Sun_Hong_Pala_Toh_Tan_Ghosal_Poria_2025}. However, several works suggest that these models exhibit limited capacity for historical and cultural reasoning, particularly in non-English cultural contexts~\cite{Hauser_Kondor_Reddish_Benam_Cioni_Villa_Bennett_Hoyer_Francois_Turchin_et_2024,Urailertprasert_Limkonchotiwat_Suwajanakorn_Nutanong_2024,Tang_Deng_Su_Yang_Wang_2024,Ghaboura_More_Thawkar_Ghallabi_Thawakar_Khan_Cholakkal_Khan_Anwer_2025,Bu_Wang_Wang_Liu_2025}.  
Moreover, even when provided with explicit historical texts, models often struggle to capture deeper cultural knowledge beyond surface-level patterns~\cite{Tang_Deng_Su_Yang_Wang_2024}, as such knowledge is structured hierarchically and exhibits diverse expressions.

Taxonomies provide a structured abstraction of domain concepts through hierarchical ``\textit{is-a}'' relations~\cite{Shang_Dash_Chowdhury_Mihindukulasooriya_Gliozzo_2020}, serving as an effective way to organize knowledge and facilitate model understanding. Compared with knowledge graphs and ontologies, taxonomies require less complex structural and semantic modeling, making them more scalable for organizing large-scale historical and cultural knowledge across long temporal spans~\cite{Franz_Thau_2010}. Moreover, taxonomies can serve as a semantic foundation for constructing knowledge graphs and ontologies~\cite{Pan_Zhang_Adamu_Dragut_Latecki_2025}, and support downstream tasks such as historical information retrieval, cultural knowledge alignment, and historically grounded reasoning in LLMs.


Previous works have explored taxonomies in specific historical domains, such as Italian public administration~\cite{pandolfo2015stole} and traditional Chinese time systems~\cite{Wang_Wang_Wei_2024}. However, historical documents are characterized by large scale, complex narratives, and highly diverse expression styles~\cite{Chang_Shiue_Yeh_Demberg_2021,Tang_Deng_Su_Yang_Wang_2024}. As a result, existing historical taxonomies remain limited in scale and cultural coverage because manual taxonomy construction is labor-intensive~\cite{Bordea_Lefever_Buitelaar_2016}. Recent studies have shown that LLMs can effectively support automated taxonomy induction and expansion~\cite{Chen_Yi_Varro_2023, lahiri-etal-2025-taxoalign, Strakova_Fucikova_Hajic_Uresova_2023, Ghamlouch_Alam_2025}, offering new possibilities for large-scale historical taxonomy construction.

Therefore, in this work, we propose a multi-agent LLM framework for historical event taxonomy construction based on the \textit{Twenty-Four Histories}, a canonical collection of official Chinese historical records spanning ancient to late imperial China. We adopt an event-centered taxonomy design, as events and their roles serve as fundamental units in historical narrative understanding~\cite{Goy_Magro_Rovera_2015, Li_Zhu_Shen_Du_Guan_Deng_2020}.
In this framework, LLM agents are assigned distinct roles across three stages. The framework first induces an initial taxonomy from raw historical corpora in a bottom-up manner, then expands it through top-down generalization using LLM world knowledge, and finally refines it by incorporating corpus-derived evidence and external structured historical resources to ensure faithfulness. This framework significantly accelerates the construction of historical knowledge organizations.

Our main contributions are summarized as follows:

\begin{itemize}
\item We propose \textbf{CHisAgent}, the multi-agent LLM framework specifically designed for large-scale historical event taxonomy induction, which decomposes the task into induction, expansion, and completion stages with explicitly role-specialized agents.

\item  We construct a large-scale event taxonomy for ancient Chinese history and culture based on the \textit{Twenty-Four Histories}, spanning over 3,000 years and covering major political, military, diplomatic, and social events.

\item We conduct extensive evaluations under both reference-free and reference-based settings, demonstrating strong coverage and alignment with source texts, and further experiments validate its effectiveness in cross-cultural alignment.

\end{itemize}

\section{Related Work}

\subsection{Automated Taxonomy Induction}

Taxonomy construction generally consists of multiple stages, including entity extraction, hierarchy induction, taxonomy expansion, and evaluation. Each stage is traditionally labor-intensive and costly, which has motivated extensive exploration into automated methods~\cite{Shen_Wu_Lei_Zhang_Ren_Vanni_Sadler_Han_2018,Chen_Lin_Klein_2021, Lee_Shen_Kang_Yoon_Han_Yu_2022}. Recent LLMs have been increasingly applied to automate multiple stages of taxonomy construction. 
For example, \citet{Zeng_Bai_Tan_Feng_Liang_Zhang_Jiang_2024} proposed an in-context learning framework (Chain-of-layer) for taxonomy induction from a given entity set, evaluated through ancestor-descendant matching against existing taxonomies using precision, recall, and F1.
\citet{Shen_Zhang_Zhang_Han_2025} reformulated entity set expansion, taxonomy expansion, and seed-guided taxonomy construction as sibling and parent identification tasks, and trained instruction-tuned LLMs to address these tasks. 
Similarly, \citet{zeng-etal-2025-codetaxo} proposed CodeTaxo, which represents taxonomies as executable Python programs for LLM-based taxonomy expansion. Their evaluation uses accuracy and Wu \& Palmer similarity to measure node assignment correctness and structural semantic similarity.

In addition, LLM-driven approaches have been applied to taxonomy construction in specific domains such as engineering, social media, and research fields~\citep{Aggarwal_Salatino_Osborne_Motta_2026, lahiri-etal-2025-taxoalign, Zhang_Zhu_Zhang_Li_2025}.

\textbf{Multi-agent framework.} Building on single-model approaches, recent studies assign LLMs specialized roles and allow them to collaborate with other agents or human experts to induce and expand taxonomies. For instance, \citet{Li_Kang_Bie_2025} proposed CLIMB, which converts clustering outputs into canonical occupational nodes and refines the hierarchy via iterative generator--evaluator interaction. Their evaluation is indirect, relying on LLM-based annotation performance to measure label consistency, coverage, and usability rather than gold-standard taxonomies.
\citet{Kargupta_Zhang_Zhang_Zhang_Mitra_Han_2025} introduced TaxoAdapt for scientific literature, a multidimensional adaptive framework that uses LLMs to adjust taxonomy width and depth based on topical distributions. It first generates a coarse taxonomy and then iteratively expands lower-level structures with LLM-guided rules. Evaluation combines automatic and human assessment of hierarchical validity, coherence, and coverage at both node and structure levels.

\subsection{Taxonomy Induction in Historical and Cultural Domains}

Prior work in cultural heritage has explored ontology-based knowledge modeling for historical artifacts. For example, a CIDOC CRM-aligned ontology~\cite{Zhang_2026} was developed for Chinese painting collections (CCACP), integrating metadata standards to support knowledge graph construction and cultural analysis.
In the context of Chinese historical knowledge, \citet{Wang_Wang_Wei_2024} proposed a time ontology for representing traditional Chinese dating systems, focusing primarily on temporal structures. 
Beyond ontologies, several studies have constructed domain-specific taxonomies for historical applications. \citet{congcong-etal-2023-ched} manually developed an event taxonomy based on the Twenty-Four Histories for event extraction, while \citet{Tang_Deng_Wang_Su_2025} designed a relation taxonomy to support structured historical relation extraction. However, existing taxonomies are relatively coarse-grained and task-oriented, with limited coverage of fine-grained conceptual distinctions. The lack of effective methods for processing large-scale historical documents has hindered further development, while the emergence of LLMs provides new opportunities for automatic knowledge organization.



\begin{figure*}
    \centering    
    \includegraphics[width=16cm,height=6.5cm]{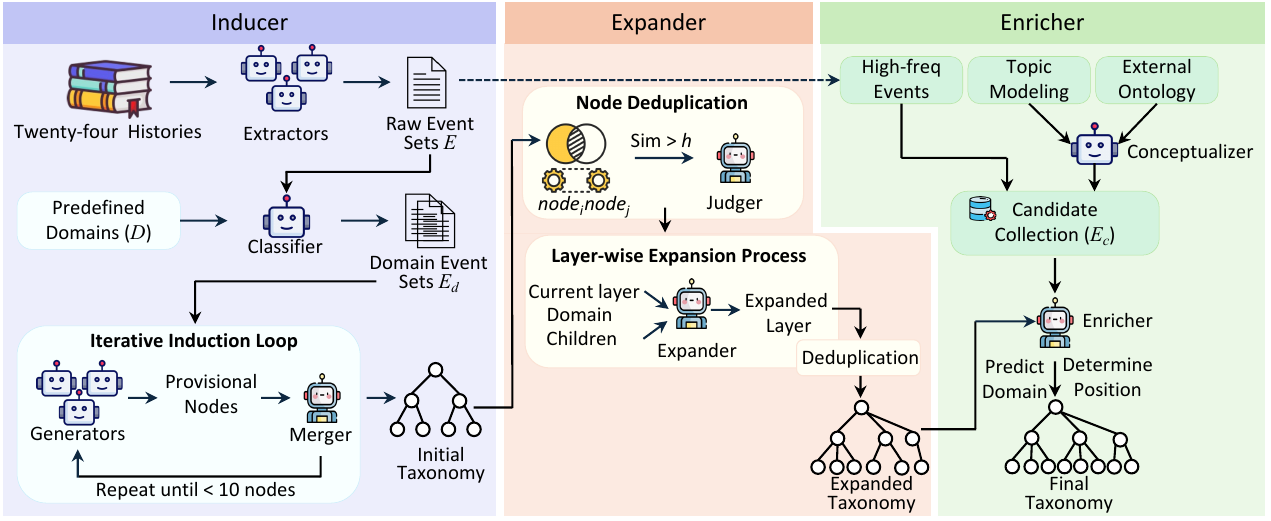}
    \vspace{-3mm}
    \caption{Illustration of the multi-agent CHisAgent framework, consisting of three key stages: the Inducer, the Expander, and the Enricher.}
    \label{framework}
    \vspace{-4mm}
\end{figure*}

\begin{figure*}
    \centering  
    \includegraphics[width=14cm,height=3cm]{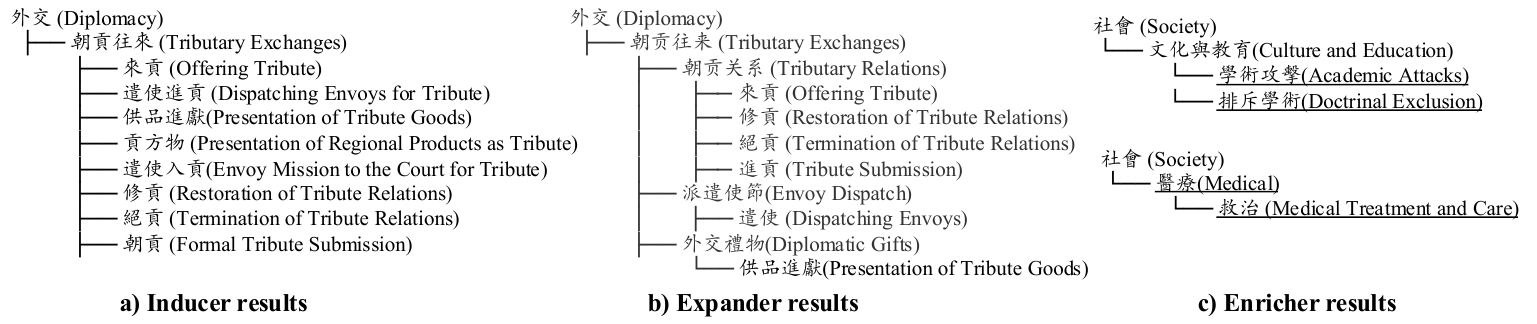}
    \vspace{-3mm}
    \caption{Illustration of different stages. The Inducer and Expander use an example from the Diplomacy domain, while the Enricher uses the Society domain. Underlined nodes are introduced by the Enricher.}
    \label{stage_case}
    \vspace{-4mm}
\end{figure*}
\section{Methodology}

Our framework consists of three LLM-based agent modules: the \textit{Inducer}, the \textit{Expander}, and the \textit{Enricher}, designed to address key challenges in historical event taxonomy construction.

Historical documents pose three major challenges in taxonomy construction. First, they often exhibit condensed expressions and stylistic variation, making event boundary detection and trigger identification difficult. Second, historical event taxonomies derived from corpora are often incomplete, with missing intermediate concepts and irregular hierarchical structures due to uneven coverage and limited structural reasoning ability. Third, historical understanding frequently requires external contextual knowledge, while LLM-generated content may suffer from semantic drift and insufficient grounding.

To address these challenges, we first introduce the \textbf{Inducer}, which performs bottom-up event hierarchy induction using a multi-extractor design with an iterative refinement loop, targeting the difficulty of event boundary detection and implicit structure extraction. However, the resulting hierarchy may remain incomplete or structurally inconsistent. Therefore, the \textbf{Expander} introduces missing intermediate concepts to improve hierarchical completeness and coherence. Finally, to ensure historical faithfulness and reduce semantic drift, the \textbf{Enricher} integrates corpus evidence with external structured historical knowledge for verification and grounding.

\subsection{Inducer}

The Inducer performs event extraction, domain classification, and bottom-up taxonomy induction in a collaborative agent framework.

We first construct the corpus from the \textit{Twenty-Four Histories}. We extract major chapter types, including ``傳'' (Biographies), ``誌'' (Treatises), ``書'' (Records), and ``紀'' (Annals), from each historical document, resulting in 192 chapters and 584,096 characters (details in Appendix~\ref{dataset}).
Then extractor agents identify event instances and event types from each chapter. We use multiple extractors to improve coverage. This yields an event set $E = \{e_1, \dots, e_M\}$. We find that LLM-based extraction is sufficiently reliable for constructing an event type pool, as further analyzed in Appendix~\ref{extractor_evaluation}.

A classifier agent assigns each event type to a domain $d \in D$ based on event descriptions and contextual evidence. Event types are aggregated within each domain as $E_d = \{e_1, \dots, e_m\}$. The domains are refined through clustering and expert validation, covering eight categories including Politics, Military, Diplomacy, Society, Ritual, Nature, and Individual.
The full description of each domain is provided in Appendix~\ref{domain_description}. The number of event types in each domain is presented in Appendix~\ref{domain_event}.

\textbf{Inducing the Initial Taxonomy.}
Within each domain, we construct the taxonomy using a bottom-up iterative process with two specialized LLM roles. Three Generator LLMs merge semantically related event types to form higher-level nodes, while a Merger LLM performs deduplication and concept unification.
The refined results are then fed back to each Generator to produce the next level of abstraction.
This iterative process continues until the number of high-level event categories within a domain falls below ten, which is empirically chosen based on preliminary experiments to balance abstraction and interpretability.
In practice, this stopping criterion is typically reached after two or three merging rounds, resulting in an initial taxonomy.

Each node is denoted as $n_{d,l,i}$, where $d$ is the domain, $l$ the hierarchy level, and $i$ the node index. The taxonomy for domain $d$ is defined as $\mathcal{T}_d = (\mathcal{N}_d, \mathcal{H}_d)$, where $\mathcal{N}_d$ denotes nodes with definitions and $\mathcal{H}_d$ denotes parent-child relations. As shown in Figure~\ref{stage_case}, we use ``Diplomacy'' as an example of the initial taxonomy produced by the Inducer (partial).

\subsection{Expander}

The Expander addresses structural incompleteness in the initial taxonomy induced by the Inducer, which is caused by loss of intermediate abstractions during bottom-up aggregation. It performs layer-aware top-down refinement to improve hierarchical completeness within each domain.

The expansion process consists of node deduplication, hierarchical expansion, and post-expansion cleaning.

\textbf{Node Deduplication.}
To reduce redundancy, we perform both within-domain and cross-domain deduplication. For any node pair $(n_i, n_j)$, we compute semantic similarity $s_{ij}=\text{Sim}(n_i,n_j)$. Nodes with $s_{ij} > h$ are considered redundant, and a judger LLM selects a representative node.

\textbf{Layer-wise Hierarchical Expansion.}
We then perform top-down expansion to recover missing or underrepresented conceptual structures. For each layer $l$, the model receives current nodes, their children, and domain context, and identifies (i) missing sibling categories, (ii) missing intermediate abstractions, and (iii) inconsistent parent–child assignments. The Expander then inserts new nodes and reassigns children to enforce hierarchical consistency. As shown in Figure~\ref{stage_case}, the Expander adds a new layer to the Diplomacy domain by inserting intermediate nodes to refine the structure of lower-level nodes.

\textbf{Post-expansion Cleaning.}
Since new nodes may introduce redundancy or semantic overlap, a final global deduplication step is applied to ensure structural compactness.

\subsection{Enricher}
The LLM-based \textbf{Enricher} improves the completeness and corpus alignment of the expanded taxonomy while preserving faithfulness to historical evidence. It follows a two-step design to recover event concepts that may be omitted during the \textit{Inducer} and \textit{Expander} stages.

\textbf{Candidate Event Collection.}
We construct a comprehensive candidate event set $E_c$ from three complementary sources. (1) \textit{High-frequency events} extracted from the initial event set $E$, where event types with frequency $f > 5$ are selected as frequent candidates ($E_{\text{freq}}$). (2) \textit{Topic-based events} derived from the corpus using BERTopic~\cite{grootendorst2022bertopic}, yielding latent topics $Topics=\{t_1,\dots,t_k\}$ that capture potential missing mid- and high-level event structures~\cite{Tang_Wang_Wang_2026}. Detailed topic statistics are provided in Appendix~\ref{Cluster_topics}. (3) \textit{Ontology-derived relations} collected from external historical knowledge bases, including the Chinese Biographical Database (CBDB) and the historical relation ontology proposed by~\citet{Tang_Deng_Wang_Su_2025}, forming a relation set $R=\{r_1,\dots,r_n\}$.

Since topics and relations are not directly aligned with event-type representations, an LLM-based \textbf{Event Conceptualizer} transforms $Topics$ and $R$ into explicit event-type candidates, producing $E_{\text{Topics}}$ and $E_{\text{R}}$. The full candidate set is:
$E_c = E_{\text{freq}} \cup E_{\text{Topics}} \cup E_{\text{R}}$.
We then perform global deduplication over $E_c$ using semantic similarity filtering to remove redundant event types.

\textbf{Positioning Candidate Events.}
For each candidate event $e \in E_c$, the Enricher first predicts its most likely domain $d$ based on event semantics and domain descriptions. It then checks redundancy with existing nodes in the current domain taxonomy $\mathcal{T}_d$. If no duplication is detected, $e$ is inserted into an appropriate position within $\mathcal{T}_d$, yielding the enriched taxonomy.

The Enricher improves corpus alignment and representational fidelity by incorporating previously missing but evidence-supported event concepts from external sources. As shown in Figure~\ref{stage_case}, it augments the taxonomy with additional nodes derived from external knowledge.

\section{Experiments}

\subsection{Experimental Setup}

Based on preliminary experiments, we assign different LLMs to specialized roles across stages according to their relative task performance.
In the extraction and induction stage, DeepSeek-V3 and Qwen3 are used as extractor agents, while DeepSeek-V3, Qwen3, and GPT-4o are used as generator agents. GPT-4o further serves as the domain classifier. GPT-5 serves as the merger role. In the expansion stage, GPT-5 functions as the judger for resolving semantic duplicates and also acts as the expander for generating higher-level taxonomy nodes.
In the enrichment stage, GPT-4o is used as the event conceptualizer, while GPT-5 serves as the Enricher to integrate candidate events into the taxonomy.
The API versions and costs of all models are provided in Appendix~\ref{cost}.

For all semantic similarity computations, taxonomy nodes and their definitions are encoded using \texttt{text-embedding-3-small}, and cosine similarity is applied for comparison. Semantic duplication is identified using a similarity threshold of 0.6, above which node pairs are considered redundant; this threshold is determined through manual evaluation.

\begin{table*}[h]
  \centering
  \small
    \caption{Evaluation results of taxonomies generated by different methods.}
  \label{tab:evaluation_metrics}
  \vspace{-3mm}
  \setlength{\tabcolsep}{0.5mm}
  \begin{tabular}{c|c|c|ccc|ccc}
    \hline
    \Xhline{1.2pt}
    \textbf{Categories} & \textbf{Methods} & \textbf{Settings} & \textbf{Path Gran.}$\uparrow$  & \textbf{CSC}$\uparrow$ & \textbf{Cover. Rate}$\uparrow$  & \textbf{Node Recall}$\uparrow$ &\textbf{Novelty}$\uparrow$ & \textbf{Significance}$\uparrow$   \\
    \hline
    \Xhline{1.2pt}
        Human&CHED & - &80.25 & 0.1432&45.39& - &-&-\\
    
    \hline
    \multirow{9}{*}{Single-agent} &Qwen3 & \multirow{3}{*}{Domain} & 95.83 &0.2566 & 35.68&43.33  &59.62&-0.0197 \\
    &DeepSeek-V3 & & 95.40 &0.1366&39.87& 43.33  &54.74&-0.0318 \\
    &GPT-4o &  & \underline{98.96} &\underline{0.2973}&30.82& 53.33 & 53.85&-0.0326 \\
    \cline{2-9}
    &Qwen3 & \multirow{3}{*}{Corpus} & 93.55 &0.1213& 46.23&48.89&  59.22&0.0174 \\
    &DeepSeek-V3 &  & 76.56 &0.1556&41.29& 42.22&44.93& -0.0211 \\
    &GPT-4o &  & \textbf{100.00} &\textbf{0.3931}& 27.47& 33.33&49.15&-0.0952\\
    \cline{2-9}
    &Qwen3 & \multirow{3}{*}{Corpus + Domain} & 88.75 & 0.2356&33.67& 36.67 &59.09&-0.0379 \\
    &DeepSeek-V3 &  & 79.80 & 0.1485&61.81&53.33&66.99&0.0769 \\
    &GPT-4o &  & 97.50 & 0.2590&37.52& 48.89 &36.36&-0.0331\\
    \hline

     \multirow{2}{*}{Non-agent} & HiExpan&\multirow{2}{*}{Corpus + Domain}&  62.80 &0.1896 & 42.21 &42.22 &74.50 &0.1098 \\
     \cline{4-9}
    & HiSetExpan  & & 83.33 & 0.2801 & 28.14 & 25.56& 71.82 &0.0515\\
    \hline
    \multirow{4}{*}{Multi-agent}&\multirow{2}{*}{CoL}& Domain & 86.39  &0.1895 &44.44  & 52.17 &\textbf{80.92}&-0.1424 \\
     & & Corpus + Domain& 80.99 &0.2479&43.72& 46.67 &50.69 & 0.0104\\
    \cline{2-9}
    &TaxoAdapt & Corpus + Domain& 70.98&0.2075 & \underline{74.17}&\underline{60.00}&\underline{69.21}&\textbf{0.3927}\\
    \cline{2-9}
     &CHisAgent&  Corpus + Domain & 91.54 & 0.2784 &\textbf{75.13}&\textbf{68.89 } &77.70&\underline{0.1887} \\
    \hline
    \Xhline{1.2pt}
  \end{tabular}
  \vspace{-2mm}
\end{table*}

\begin{table*}[h]
  \centering
  \small
    \caption{Ablation results of different components in our framework.}
  \label{ablation}
  \vspace{-4mm}
  \setlength{\tabcolsep}{0.8mm}
  \begin{tabular}{c|c|c|c|ccc|ccc}
    \hline
    \Xhline{1.2pt}
    \multirow{2}{*}{\textbf{Method}} & \multicolumn{3}{c|}{\textbf{Components}} & \multirow{2}{*}{\textbf{Path Gran.}}&  \multirow{2}{*}{\textbf{CSC}} &\multirow{2}{*}{\textbf{Cover. Rate}} &\multirow{2}{*}{\textbf{Node Recall}}  &\multirow{2}{*}{\textbf{Novelty}} & \multirow{2}{*}{\textbf{Significance}}   \\
    \cline{2-4}
    &\textbf{Inducer} & \textbf{Expander} & \textbf{Enricher}&&&&&\\
    \hline
    \multirow{3}{*}{CHisAgent} & $\checkmark$ & $\times$ & $\times$ & 90.45 & 0.2761 &71.61& 61.11  &77.22&0.1310\\
   & $\checkmark$ & $\checkmark$ & $\times$ &  94.35  & 0.2464& 72.36& 65.56 & 77.59& 0.1667  \\
   & $\checkmark$ & $\checkmark$ & $\checkmark$ & 91.54 &  0.2784 &75.13 & 68.89 &77.70& 0.1887\\
    \hline
    \Xhline{1.2pt}
  \end{tabular}
  \vspace{-3mm}
\end{table*}

\subsection{Baselines}

We compare our framework with four categories of taxonomy construction methods.

\textbf{Traditional Taxonomy Induction Methods.}
We reproduce two representative non-LLM taxonomy induction approaches: HiExpan and HiSetExpan~\cite{Shen_Wu_Lei_Zhang_Ren_Vanni_Sadler_Han_2018}.

\textbf{Single-Agent LLM Methods.}
We design three prompting-based taxonomy construction settings using GPT-4o, DeepSeek-V3, and Qwen3:

\textbf{LLM + Domains}, where the model generates taxonomies from predefined domains in a top--down manner;

\textbf{LLM + Event Corpus}, where the model induces taxonomies bottom--up from fine-grained event types;

\textbf{LLM + Event Corpus + Domains}, where both domain knowledge and event types are provided for intermediate taxonomy induction.

\textbf{Multi-Agent Methods.}
We further compare with two recent multi-agent taxonomy construction frameworks: Chain-of-Layer (CoL)~\cite{Zeng_Bai_Tan_Feng_Liang_Zhang_Jiang_2024} and TaxoAdapt~\cite{Kargupta_Zhang_Zhang_Zhang_Mitra_Han_2025}.

\textbf{Human-Crafted Taxonomy.}
We include CHED~\cite{congcong-etal-2023-ched} as a human-authored reference taxonomy derived from the \textit{Twenty-Four Histories}.

Implementation details of all baselines are provided in Appendix~\ref{Baselines}.

\subsection{Evaluation Metrics}

We evaluate the constructed taxonomies using two types of metrics, reference-free and reference-based, covering structural and semantic aspects. The detailed calculation procedures for all metrics are provided in Appendix~\ref{metrics}. 

\textbf{Reference-free Metrics} are as follows.

\textbf{Path Granularity}~\cite{Kargupta_Zhang_Zhang_Zhang_Mitra_Han_2025}. This metric assesses whether a node ($n_{d,l,i}$) is semantically more specific than its parent. The score (0 or 1) is assigned by GPT-4o.

\textbf{CSC}~\cite{Wullschleger_Zarharan_Daly_Pouly_Foster_2025}. A structural consistency metric that evaluates whether distances between nodes align with their semantic relations. Higher scores indicate stronger alignment.

\textbf{Coverage Rate}. This metric measures coverage of historical events in the \textit{Twenty-Four Histories}, extracted from non-overlapping chapters.

\textbf{Reference-Based Metrics.}
These metrics compare the constructed taxonomy with the human-curated taxonomy of CHED~\cite{congcong-etal-2023-ched}, capturing different aspects of deviation from the reference.

\textbf{Node Recall}. Measures the proportion of human taxonomy nodes captured by the constructed taxonomy under a semantic similarity threshold.

\textbf{Novelty}~\cite{Kaplan_Kuhn_Hahner_Benkler_Keim_Fuchb_Corallo_Heinrich_2022}. Quantifies the proportion of newly introduced nodes not present in the human taxonomy. Higher values may indicate either true extensions or over-generation beyond the reference.

\textbf{Significance}~\cite{Kaplan_Kuhn_Hahner_Benkler_Keim_Fuchb_Corallo_Heinrich_2022}. Measures the difference in granularity compared to the human taxonomy. Positive values indicate finer-grained structures, while negative values indicate coarser ones.

We note that reference-based metrics should be interpreted jointly, as they reflect different aspects of coverage and deviation from the human taxonomy rather than uniformly better or worse quality.

\subsection{Overall Experimental Results}
The evaluation results are presented in Table~\ref{tab:evaluation_metrics}. We first observe that single-agent generation methods tend to achieve higher Path Granularity and CSC scores than multi-agent methods, as they construct the entire taxonomy in a unified process. Traditional taxonomy induction methods show no clear advantage in complex systems with large-scale event type spaces.
In contrast, multi-agent methods provide a more explicit decomposition of the taxonomy construction process, leading to improved structural coverage and better alignment with reference taxonomies. In particular, our method achieves the best results in terms of Coverage Rate and Node Recall. TaxoAdapt obtains the highest Significance score, indicating that it produces the most fine-grained taxonomy relative to the human reference. Chain-of-Layer (Domain) achieves the best performance on the Novelty metric.
Statistics of the generated taxonomies are provided in Appendix~\ref{taxonomy_statis}.

\subsection{Ablation Study}
Table~\ref{ablation} presents an ablation study evaluating the contributions of the three agent modules: Inducer, Expander, and Enricher.
Using only the Inducer yields the lowest Node Recall and Coverage Rate, indicating that the induced taxonomy remains relatively coarse-grained and incomplete. Adding the Expander substantially improves Path Granularity and Node Recall, showing its effectiveness in constructing deeper hierarchical structures. The slight decrease in CSC suggests that maintaining semantic consistency becomes more challenging as the taxonomy expands. 
Further incorporating the Enricher improves Coverage Rate, Node Recall, and Significance, indicating that it effectively supplements missing concepts and enriches historical content. Overall, the results show that the three modules play complementary roles in improving taxonomy quality.

\subsection{Domain Analysis}

\begin{figure}[ht]
    \centering
        \subfigure[Coverage Rate.] {
     \label{f2b}     
\includegraphics[width=0.45\columnwidth]{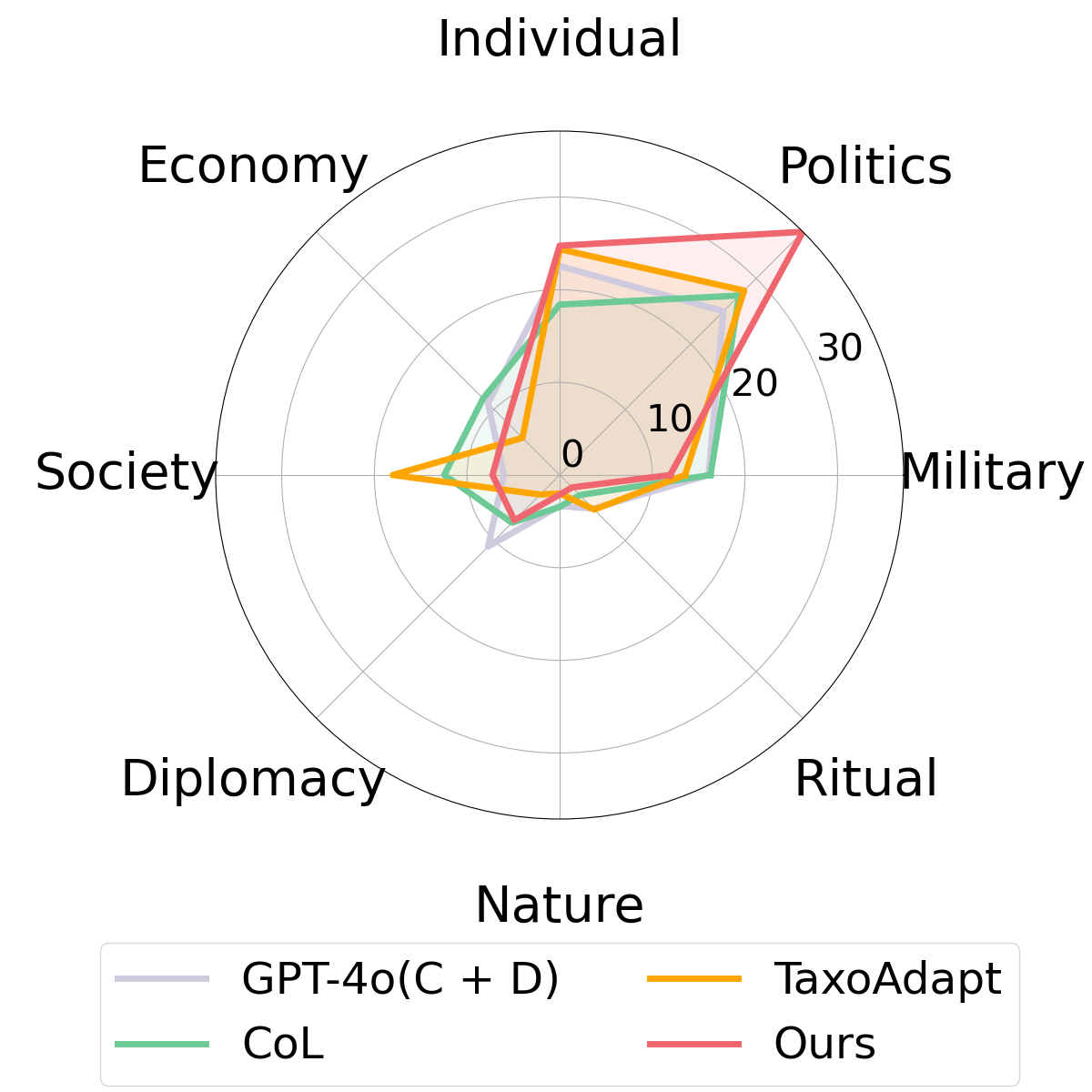}  
    } 
    \subfigure[Novelty.] {
    \label{f2d}     
    \includegraphics[width=0.45\columnwidth]{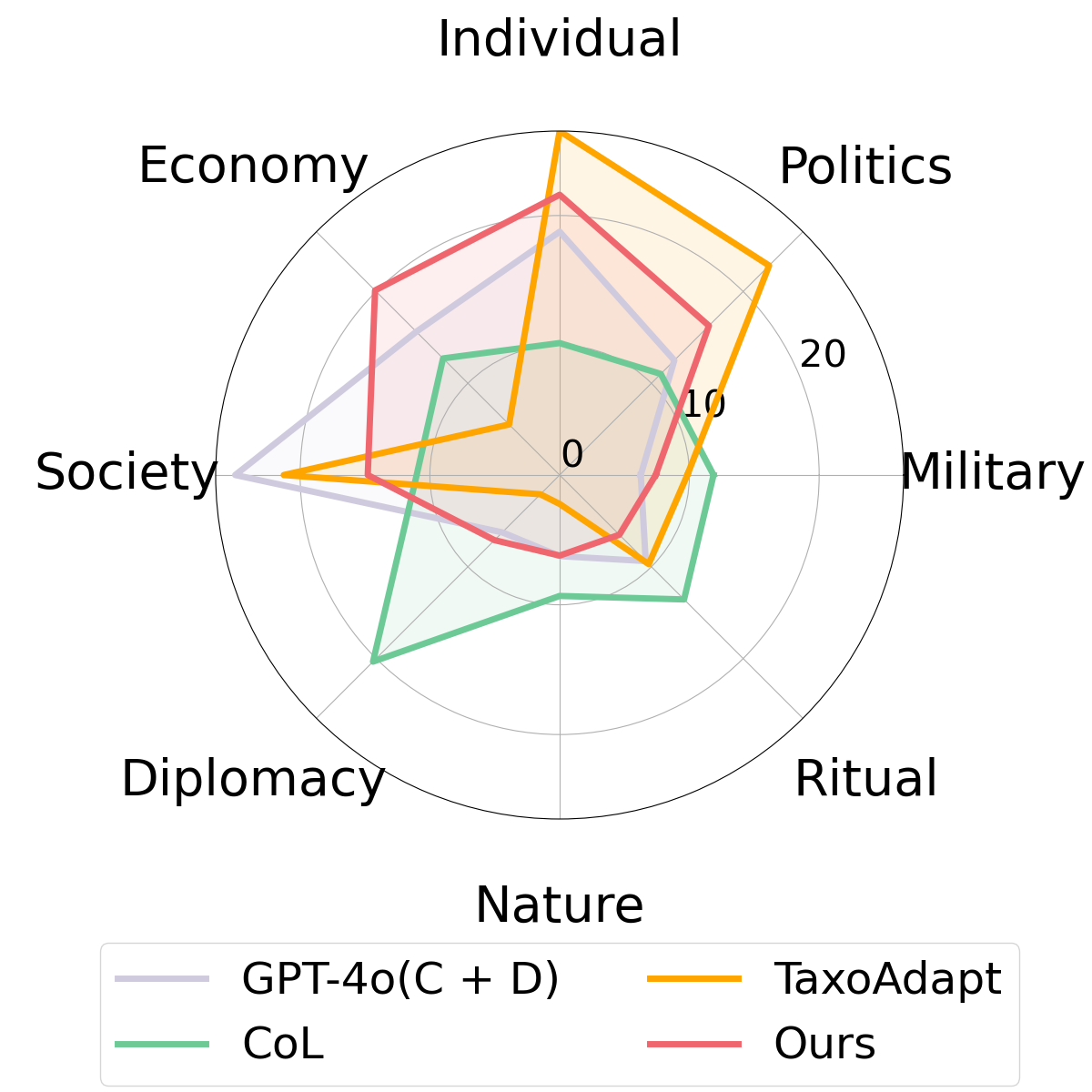}  }
    \vspace{-3mm}
    \caption{Radar chart of evaluation metrics of different methods across various domains. GPT-4o (C + D) denotes GPT-4o (Corpus + Domain), and ``Economy'' represents ``Economy-Livelihood''.}
    \label{f2}
     \vspace{-6mm}
\end{figure}

We further compare taxonomies generated by GPT-4o (Corpus + Domain), CoL (Corpus + Domain), TaxoAdapt, and our method across eight domains using the radar charts in Figure~\ref{f2}. 

Figure~\ref{f2b} reports Coverage Rate, measuring how well each method captures event types extracted from other chapters of the \textit{Twenty-Four Histories}. Most methods obtain higher coverage in Politics, Military, and Individual categories, while TaxoAdapt additionally performs well in Society. This metric is partially influenced by the quality and recall of LLM-based event extraction, which may introduce domain-specific biases.

In addition, Figure~\ref{f2d} reports Novelty, reflecting the introduction of nodes absent from the human taxonomy. All methods generate the largest number of novel nodes in Individual, Society, and Politics. Our method shows particularly high novelty in Individual and Economy-Livelihood, whereas GPT-4o and TaxoAdapt produce more novel nodes in Society. In contrast, CoL exhibits a relatively balanced novelty distribution across domains.

Additional analyses of parent--child relationship rationality and recalled node distributions are provided in Appendix~\ref{appendix_domain_analysis}.

\begin{figure*}
\centering\includegraphics[width=0.8\linewidth]{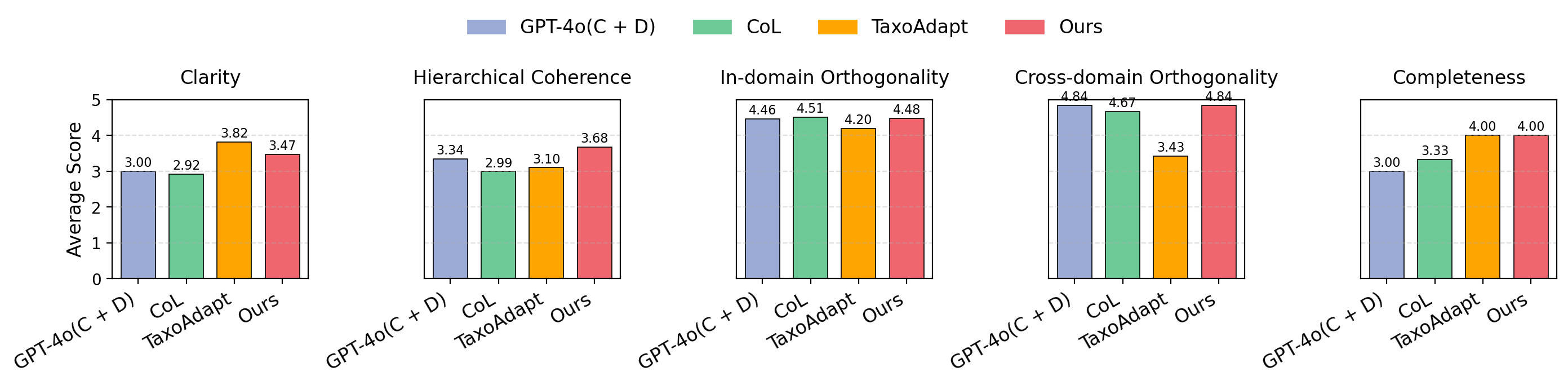}
    \vspace{-3mm}
    \caption{Average human evaluation scores of taxonomies generated by different methods across five dimensions.}
    \label{human_score}
    \vspace{-5mm}
\end{figure*}

\subsection{Human Evaluation}
\label{human_eval}
Following \citet{Zhang_Zhu_Zhang_Li_2025}, we conduct a human evaluation to assess the quality of taxonomies generated by different methods (GPT-4o (Corpus + Domain), CoL (Corpus + Domain), TaxoAdapt, and Ours).
The evaluation focuses on semantic clarity, structural coherence, and semantic distinction, and considers five complementary perspectives as follows.

\textbf{Node Clarity.}
Evaluators judge whether the node’s semantics are clear and whether the node can appropriately represent an event type in ancient Chinese cultural system. Scores range from 0 to 5.

\textbf{Hierarchical Coherence.}
Evaluators assess whether the hierarchical structure from the root to the leaf is logically coherent. Scores range from 0 to 5.

\textbf{In-domain Orthogonality.}
Evaluators measure the degree of semantic independence among nodes within the same domain, assessing whether taxonomy concepts exhibit minimal semantic overlap and redundancy. Scores range from 0 to 5.

\textbf{Cross-domain Orthogonality.}
Evaluators assess whether nodes assigned to different domains are semantically distinguishable from each other. Scores range from 0 to 5.

\textbf{Completeness.}
Evaluators judge whether each complete taxonomy sufficiently covers the conceptual space of Chinese historical events. Scores range from 0 to 5.

Details of node sampling and evaluation guidelines are provided in Appendix~\ref{guideline}. 

We employ three independent evaluators with expertise in Chinese history. After two rounds of discussion, the inter-rater agreement demonstrates moderate-to-substantial consistency across evaluators, with a weighted Cohen’s kappa of 0.6982, a Spearman correlation of 0.5722, and an ICC(2,1) score of 0.6992, indicating generally reliable human judgments. Following prior work, the final scores were obtained by averaging annotator ratings.
Human evaluation results are presented in Figure~\ref{human_score}. TaxoAdapt achieves the highest scores in Clarity and Completeness, but obtains relatively lower Cross-domain Orthogonality, suggesting semantic overlap across domains despite extensive node generation. In contrast, GPT-4o (Corpus + Domain) attains the highest Cross-domain Orthogonality, indicating strong separation between categories, but exhibits lower completeness.
Overall, our method achieves a more balanced performance across all metrics, maintaining high levels of completeness, semantic clarity, and structural orthogonality, which aligns closely with human judgments.
A more detailed analysis of model error patterns is provided in section~\ref{error_analysis}.

\subsection{Human Evaluation of Event Extraction}

Since extraction errors may propagate to subsequent taxonomy construction stages, we further evaluate the quality of the Extractor agents through a targeted human evaluation. We evaluate the event extraction results of Qwen3 and DeepSeek-V3 on 10 Classical Chinese historical passages. Two annotators independently judged whether each extracted event was correct and identified events missed by the Extractor agents. Due to the limited availability of large-scale annotated document-level event extraction datasets for Classical Chinese historical texts, we adopt human evaluation rather than directly comparing with existing document-level event extraction methods that require task-specific training data.

The two annotators achieved an overall agreement of 0.83 on event correctness. To provide a conservative evaluation, an extracted event is considered correct only when both annotators agree that it is valid; otherwise, it is counted as an extraction error. Table~\ref{tab:event_extraction_human} summarizes the results. Both extractors reliably identify major historical events. Qwen3 achieves broader coverage with higher recall (0.6588), whereas DeepSeek-V3 adopts a more conservative extraction strategy, achieving higher precision (0.8312). Overall, the two extractors obtain comparable F1 scores.

\begin{table}[t]
\centering
\small
\caption{Human evaluation results for event extraction by different Extractor agents.}
\label{tab:event_extraction_human}
\setlength{\tabcolsep}{0.8mm}
\begin{tabular}{lccccc}
\toprule
Extractor & Extracted & Correct & Missed & Precision & Recall \\
\midrule
Qwen3 & 242 & 195 & 101 & 0.8058 & 0.6588 \\
DeepSeek-V3 & 237 & 197 & 120 & 0.8312 & 0.6215 \\
\bottomrule
\end{tabular}
\end{table}

We further analyze the missed and incorrect extractions. Most missed cases involve fine-grained sub-events within long sentences rather than completely overlooked events. The main sources of missed events include: (1) \textit{granularity differences}, where models extract a main event while annotators identify additional fine-grained actions; (2) \textit{multi-event sentence structures}, where multiple coordinated actions are present but models identify only the dominant event; and (3) \textit{implicit event expressions}, which require contextual inference rather than explicit trigger words.

False positives mainly arise from overly broad interpretations of frequent trigger words, such as 賜'' (grant/bestow), 聞’’ (hear/learn of), 誅'' (execute/punish), 遣’’ (send/dispatch), and ``曰’’ (say/state), as well as from treating background or consequence descriptions as independent events. These errors highlight the challenges inherent in Classical Chinese historical texts, including ambiguous event boundaries, ellipsis, and complex narrative structures.

\subsection{Robustness of LLM-based Evaluation}

To examine whether the evaluation results are sensitive to the choice of LLM evaluator, we conduct an additional robustness analysis using GLM-5.2 ($T=0$), which belongs to a different model family from the GPT models used in our original evaluation. We evaluate Path Granularity using the same evaluation protocol and prompts as in the original experiments. The results are shown in Table~\ref{tab:evaluator_robustness}.

Although the absolute scores vary somewhat across evaluators, the relative performance patterns remain largely consistent. In particular, GPT-4o (Corpus) and CHisAgent achieve the highest scores, while HiExpan and TaxoAdapt obtain lower scores. This consistency across different model families suggests that the comparative conclusions are not strongly dependent on a specific LLM evaluator or its evaluation preferences.

\begin{table}[t]
\centering
\small
\caption{Robustness analysis of Path Granularity evaluation using different LLM evaluators.}
\label{tab:evaluator_robustness}
\begin{tabular}{lcc}
\toprule
Methods & GPT-4o & GLM-5.2 \\
\midrule
GPT-4o (Corpus) & 100.00 & 98.28 \\
HiExpan (Corpus+Domain) & 62.80 & 57.81 \\
HisetExpan (Corpus+Domain) & 83.33 & 83.33 \\
CoL (Corpus+Domain) & 80.99 & 71.53 \\
TaxoAdapt (Corpus+Domain) & 70.98 & 70.62 \\
CHisAgent (Corpus+Domain) & 91.54 & 82.67 \\
\bottomrule
\end{tabular}
\end{table}

This cross-family evaluation provides additional evidence that our evaluation findings are robust to evaluator choice, mitigating concerns about potential evaluator bias arising from using models from the same family.

\subsection{Cross-Cultural Generalization across East Asian Historical Corpora}

 In this section, we evaluate whether the induced taxonomies generalize across historical corpora within the East Asian cultural sphere.
We consider representative records from Japan, Korea, and Vietnam, including the \textit{Nihon Shoki} (日本書紀)\footnote{https://en.wikipedia.org/wiki/Nihon\_Shoki}, the \textit{Joseon Sillok} (朝鮮實錄)\footnote{https://hanchi.ihp.sinica.edu.tw/mqlc/hanjishilu}, and the \textit{Dai Viet Su Ky Toan Thu} (大越史記全書)\footnote{https://ctext.org/wiki}.

For each corpus, we randomly sample several chapters and extract events using Qwen3 and DeepSeek-V3. As all corpora are written in classical Chinese, we assume comparable extraction quality across datasets. The resulting number of event types in each corpus is approximately 300–400.
We report the Coverage Rate of each taxonomy on these corpora in Table~\ref{cross_culture}.
With the exception of CHED, all methods achieve their lowest coverage on the Vietnamese corpus, suggesting greater divergence from Chinese-centered taxonomies than the Korean and Japanese records.

\textbf{Case Study.}
We further conduct a fine-grained analysis of tributary interactions as a representative diplomatic event type.
Although tribute relations appear across East Asian polities, the induced taxonomy reveals systematic differences in their narrative organization.
Specifically, as shown in Figure~\ref{stage_case}, tribute events are mapped to separate nodes for incoming tribute (\emph{來貢}) and outgoing tribute (\emph{進貢}), capturing different narrative perspectives.

We manually verify all extracted tribute-related event samples in each corpus. Then we observe that events from Chinese and Japanese corpora predominantly align with the incoming tribute (\emph{來貢}) perspective, Vietnamese records more frequently align with outgoing tribute (\emph{進貢}), and Korean records exhibit a mixed pattern.
These findings demonstrate the importance of perspective-aware taxonomy nodes for cross-cultural comparison, allowing semantically similar practices to be aligned while preserving differences in historical narration.

\subsection{Error Analysis}
\label{error_analysis}

We conduct a quantitative error analysis based on the human evaluation results (Section~\ref{human_eval}). Among the sampled nodes, 32\% received low scores (lower than 4) for hierarchical coherence, making structural inconsistency the dominant source of errors. This suggests that the most common error type is hierarchical over-segmentation, where the model introduces near-synonymous intermediate nodes without clear conceptual distinctions. Examples include ``任用 (Appointment) $\rightarrow$ 任命 (Nomination)'' and ``辯解 (Justification) $\rightarrow$ 解釋 (Explanation)''. These errors suggest that the model may confuse lexical similarity with true hierarchical subsumption, leading to overly fine-grained or weakly differentiated taxonomic structures.
Other error types are provided in Appendix~\ref{appendix_error_analysis}.

\begin{table}[t]
  \small
  \centering
      \caption{Cross-corpus Coverage Rate of taxonomies generated by different methods.}
  \label{cross_culture}
    \vspace{-3mm}
  \setlength{\tabcolsep}{1.5mm}
  \begin{tabular}{c|c|c|c}
    \hline
    \Xhline{1.2pt}
    Methods& Korea & Japan & Vietnam \\
    \hline
    CHED&43.73 &53.90 & 50.42\\
    CoL&29.74 & 29.96 & 18.78\\
    TaxoAdapt & 75.13 & 74.04& 56.71\\
    CHisAgent & 75.51 &80.50 & 60.55\\
    \hline
    \Xhline{1.2pt}
  \end{tabular}
  \vspace{-3mm}
\end{table}

\section{Conclusion}

In this study, we propose CHisAgent, a multi-agent framework for constructing a large-scale taxonomy of ancient Chinese historical events. The taxonomy construction process is decomposed into three stages, Inducer, Expander, and Enricher, implemented by specialized LLM agents for hierarchical induction, semantic expansion, and taxonomy refinement. Experimental results demonstrate that our framework produces taxonomies with strong semantic clarity, hierarchical coherence, and broad coverage across multiple historical domains. We further show that the induced taxonomy generalizes to East Asian historical corpora. It supports cross-cultural alignment of semantically related historical events, while preserving differences in narrative perspective.
Beyond taxonomy construction, our work highlights the potential of structured event taxonomies for historical knowledge organization and AI-assisted cultural understanding. In future work, we plan to integrate the taxonomy into a historical event retrieval system to facilitate more accurate event exploration and knowledge access.
Our analysis also reveals several remaining challenges, including hierarchical over-segmentation, mixed ontological criteria, and cross-domain semantic overlap. Future work will explore ontology refinement under stronger logical constraints, temporal and causal event modeling, and validation across multilingual historical corpora.

\section*{Limitations}
This work has several limitations.

First, although we employ multiple LLMs to reduce the bias introduced by any single model, LLM-based taxonomy construction inevitably inherits biases from model pretraining. In particular, modern language models may project contemporary conceptual structures onto historical contexts or introduce hallucinated abstractions when expanding sparse historical evidence.

Second, the performance of current LLMs on Classical Chinese remains limited. Classical Chinese texts are highly concise, context-dependent, and often ambiguous, posing challenges for accurate semantic interpretation. As a result, the generated taxonomy may still contain errors or omissions, and human intervention is occasionally required to ensure correctness.

Third, while the \textit{Twenty-Four Histories} constitute a highly authoritative and representative corpus of ancient Chinese historiography, they primarily reflect official and elite perspectives. Certain aspects of ancient Chinese culture, such as folk practices, regional traditions, and informal social dynamics, are underrepresented, limiting the completeness of the resulting taxonomy.

Fourth, our event-centered taxonomy design, while effective for modeling historical narratives and social processes, may not fully capture cultural phenomena that evolve gradually or are not naturally event-driven, such as philosophical traditions or long-term ideological shifts.

Fifth, our framework relies on LLM-based agents for hierarchical induction, expansion, and refinement, which introduces inherent stochasticity into the construction process. As a result, running the framework multiple times may yield slightly different taxonomy structures. While our experiments show that the overall high-level organization and major event categories remain stable across runs, variations may still occur in fine-grained node formation and hierarchical branching. A systematic study of structural variance and stability across multiple runs is left for future work.

Finally, although we employ both automatic metrics and human evaluation, these evaluation protocols have inherent limitations. Structural metrics cannot fully capture cultural appropriateness, and human evaluation is constrained in scale and may be influenced by annotators’ backgrounds. 


\section*{Ethics Statement}
The human evaluations in this study were conducted by internal members of the research team (excluding the authors). All participants are research assistants and graduate students with expertise in the relevant domain. Participation was entirely voluntary, and all participants were informed of the study’s objectives and the nature of the tasks. No personal, sensitive, or identifiable information was collected during the evaluation.

Beyond the human evaluation process, our work aims to facilitate the organization and exploration of historical knowledge through automated taxonomy construction. However, automatically generated taxonomies may amplify biases present in historical sources or models, oversimplify complex historical events, or be misused without appropriate expert verification. To mitigate these risks, we emphasize that the generated taxonomies are intended as research aids and should not be treated as authoritative historical interpretations or replacements for expert judgment.

\section*{Acknowledgments}

This work was supported by the General Research Fund (GRF) of the Research Grants Council (RGC) of Hong Kong (Grant No. 15609725), the National Natural Science Foundation of China (Grant No. 72204258), and The Hong Kong Polytechnic University under the project ``Evaluating the Syntax-Semantics Knowledge in Large Language Models'' (Grant No. P0055270).

\bibliography{custom}
\clearpage
\appendix

\section{Dataset}
\label{dataset}
We select the version of \textit{Twenty-Four Histories} are published by Zhonghua Book Company (中華書局). Table~\ref{tab:book_stats} reports the character counts of sampled chapters from each book in the \textit{Twenty-Four Histories}, which serve as a proxy for corpus size across historical periods. We open all data at~\url{https://anonymous.4open.science/r/ACL-BC1E/}.

\section{Extractor Evaluation}
\label{extractor_evaluation}

We randomly sample 400 event instances from the outputs of the extractors and manually evaluate the extraction performance. The results show that the extraction achieves an accuracy of 85\%. 

The most common errors include non-event summaries (e.g., entity or identity descriptions), behavioral attribute descriptions, and cases where multiple events are merged into a single expression (e.g., “A and B” format), which is not suitable for higher-level event abstraction. Overall, factual extraction errors account for less than 5\% of the total samples.

\section{Domain Description}
\label{domain_description}
\begin{itemize}
    \item \textit{\textbf{Politics}: Encompasses events related to state governance, power structures, policy formulation and implementation, as well as judicial and legal affairs.}

    \item \textit{\textbf{Military}: Encompasses events related to warfare, military operations, strategic planning, and army organization and management.}

    \item \textit{\textbf{Diplomacy}: Includes inter-state political and cultural interactions conducted through official channels, such as envoy missions, tribute relations, and peace negotiations.}

    \item \textit{\textbf{Society}: Covers social activities, cultural practices, customs, and collective events involving groups, organizations, or institutions beyond the individual level.}

    \item \textit{\textbf{Ritual}: Refers to ceremonial and ritual activities conducted by ancient states or royal families.}

    \item \textit{\textbf{Economy--Livelihood}: Covers events related to economic production, resource distribution, daily subsistence, and state finance.}

    \item \textit{\textbf{Nature}: Includes natural phenomena, related events, and their impacts on human society.}

    \item \textit{\textbf{Individual}: Covers events related to individual life trajectories, identity attributes, personal behaviors, and social relationships.}
\end{itemize}

\begin{table}[h]
    \centering
    \small
        \caption{Character statistics of sampled chapters from the \textit{Twenty-Four Histories}.}
    \label{tab:book_stats}
    \setlength{\tabcolsep}{0.1mm}
    \begin{tabular}{l|c}
        \hline
        \textbf{Historical Book} & \textbf{Characters} \\
        \hline
        \textit{Records of the Grand Historian} (史記) & 39,267 \\
        \textit{Book of Han} (漢書) & 37,100 \\
        \textit{Book of Later Han} (後漢書) & 26,039 \\
        \textit{Records of the Three Kingdoms} (三國志) & 14,170 \\
        \textit{Book of Jin} (晉書) & 32,849 \\
        \textit{Book of Song} (宋書) & 46,604 \\
        \textit{Book of Southern Qi} (南齊書) & 25,738 \\
        \textit{Book of Liang} (梁書) & 37,690 \\
        \textit{Book of Chen} (陳書) & 20,338 \\
        \textit{Book of Wei} (魏書) & 29,581 \\
        \textit{Book of Northern Qi} (北齊書) & 11,885 \\
        \textit{Book of Zhou} (周書) & 41,639 \\
        \textit{Book of Sui} (隋書) & 31,472 \\
        \textit{History of the Southern Dynasties} (南史)&9146 \\
        \textit{History of the Northern Dynasties} (北史)&11311\\
        \textit{Old Book of Tang} (舊唐書) & 32,175 \\
        \textit{New Book of Tang} (新唐書) & 19,132 \\
        \textit{Old History of the Five Dynasties} (舊五代史) & 14,405 \\
        \textit{New History of the Five Dynasties} (新五代史) & 10,845 \\
        \textit{History of Song} (宋史) & 17,861 \\
        \textit{History of Liao} (遼史) & 14,869 \\
        \textit{History of Jin} (金史) & 23,591 \\
        \textit{History of Yuan} (元史) & 18,297 \\
        \textit{History of Ming} (明史) & 18,092 \\
        Total & 584,096 \\
        \hline
    \end{tabular}
\end{table}

\section{Event Type Distribution Across Domains}
\label{domain_event}

We report the event type distribution across domains in Table~\ref{statistics_event_type}.
\begin{table}[h]
    \centering
    \small
        \caption{Distribution of extracted event types across different domains.}
    \label{statistics_event_type}
    \begin{tabular}{l|c}
        \hline
        \textbf{Domain} & \textbf{Number of Event Types} \\
        \hline
        Politics & 1681 \\
        Military & 690 \\
        Diplomacy & 168 \\
        Society & 710 \\
        Ritual & 202 \\
        Economy--Livelihood & 283 \\
        Nature & 119 \\
        Individual & 1023 \\
        \hline
    \end{tabular}
\end{table}

\section{Clustering Topics}
\label{Cluster_topics}

Following prior work by \citet{Tang_Wang_Wang_2026}, which manually constructs a relation ontology for classical Chinese documents using document-derived topics as the organizational basis, we leverage topic clustering from the \textit{Twenty-Four Histories} to support taxonomy construction.

We first segment all documents in the corpus using an ancient Chinese word segmenter\footnote{\url{https://github.com/tangxuemei1995/Ancient-chinese-segmenter}}. Each paragraph is then treated as a basic textual unit and encoded with BERTopic. Based on multiple experimental trials and manual inspection, we set the number of topics to 30.
For each topic, we extract the top 30 representative words. These words are subsequently summarized into a single concept word or phrase using GPT-4o, yielding a compact semantic label for each topic.
Finally, we apply human validation to ensure that each topic label is semantically coherent and appropriate for the historical context, resulting in 23 retained topics (Table~\ref{topics}).

\begin{table}[t]
\centering
\small
\caption{Clustered historical topics extracted from the \textit{Twenty-Four Histories}.}
\label{topics}
\begin{tabular}{cl}
\hline
\textbf{Topic ID} & \textbf{Topic Label} \\
\hline
0  & 戰爭行動 (Military Actions) \\
1  & 節令時序 (Seasonal Calendar) \\
2  & 政務管理 (Administrative Governance) \\
3  & 詔令文書 (Imperial Decrees) \\
4  & 晉升調任 (Promotion and Transfer) \\
5  & 天文觀測 (Astronomical Observation) \\
6  & 皇室活動 (Royal Activities) \\
7  & 外交朝貢 (Diplomacy and Tribute) \\
8  & 地理 (Geography) \\
9  & 紀年 (Chronology) \\
10 & 自然災害 (Natural Disasters) \\
11 & 科舉考試 (Imperial Examinations) \\
12 & 言行紀錄 (Speech and Deeds) \\
13 & 登基繼位 (Accession and Enthronement) \\
14 & 年號 (Reign Titles) \\
15 & 音樂樂器 (Music and Instruments) \\
16 & 干支 (Ganzhi Cycles) \\
17 & 赦免公告 (Amnesty Decrees) \\
18 & 祭祀禮儀 (Rituals and Sacrifices) \\
19 & 帝王稱謂 (Imperial Titles) \\
20 & 邊疆防務 (Frontier Defense) \\
21 & 官員任命 (Official Appointments) \\
22 & 巡察 (Inspection) \\
\hline
\end{tabular}
\end{table}

\section{API Version and Cost}
\label{cost}

We use the following model versions in our experiments:
GPT-4o (\texttt{gpt-4o-2024-11-20}),
DeepSeek-V3 (\texttt{DeepSeek-V3-0324}),
Qwen3 (\texttt{qwen-plus-2025-07-28}),
and GPT-5 (\texttt{gpt-5-2025-08-07}).
For all models except GPT-5, the temperature is set to 0. 

The majority of token consumption is attributed to Qwen3, DeepSeek-V3, and GPT-4o, which are primarily used in the extractor stage for processing historical corpora and generating event sentences, event types, and trigger words.

In contrast, GPT-5 is mainly used in the Expander and Enricher stages, which involve significantly fewer tokens due to their more structured generation process.

The total API cost of the entire pipeline is approximately \$100.

\section{Baselines}
\label{Baselines}

We compare our framework with representative taxonomy construction methods spanning traditional non-agentic approaches, single-agent LLM prompting settings, multi-agent frameworks, and human-crafted taxonomies.

\textbf{Traditional Non-Agentic Methods.}
We reproduce two representative taxonomy induction methods: HiExpan~\cite{Shen_Wu_Lei_Zhang_Ren_Vanni_Sadler_Han_2018} and HiSetExpan~\cite{Shen_Wu_Lei_Zhang_Ren_Vanni_Sadler_Han_2018}. 

For HiExpan, we construct domain-specific seed taxonomies from our taxonomy and iteratively expand taxonomy hierarchies through both depth expansion and width expansion. Width expansion enlarges sibling sets using skip-patterns, embedding similarity, and sibling consistency, while depth expansion discovers child concepts under existing taxonomy nodes. Compared with the original setting, we adapt the method to Classical Chinese corpora and replace REPEL with BGE-M3 embeddings.

HiSetExpan is a baseline method proposed by ~\citet{Shen_Wu_Lei_Zhang_Ren_Vanni_Sadler_Han_2018}. HSetExpan differs from HiExpan in that it iteratively applies SetExpan~\cite{Shen_Wu_Lei_Shang_Ren_Han_2019} at each taxonomy level. For each lower-level node, the method attaches it to the parent node with the highest parent--child similarity score.

\textbf{Single-Agent LLM Methods.}
We design three prompting-based taxonomy construction settings using GPT-4o, DeepSeek-V3, and Qwen3.

\begin{itemize}
    \item  \textbf{LLM + Domains.}
The model is given only predefined domains and generates the taxonomy in a top--down manner by progressively producing fine-grained event types.

    \item  \textbf{LLM + Event Corpus.}
The model is provided with fine-grained event types extracted from the corpus and induces the taxonomy bottom--up through iterative clustering and abstraction.

\item  \textbf{LLM + Event Corpus + Domains.} The model is provided with both predefined domains and event-type corpus, and induces intermediate taxonomy levels by integrating domain knowledge with event types.
\end{itemize}

\textbf{Multi-Agent Methods.}
We further compare with two recent multi-agent taxonomy construction frameworks.

\textbf{Chain-of-Layer}~\cite{Zeng_Bai_Tan_Feng_Liang_Zhang_Jiang_2024} constructs taxonomies in a top--down manner through iterative layer-wise entity selection. Following the original framework, we first build a candidate event-type pool from automatically generated event inventories, then iteratively generate and select intermediate taxonomy nodes layer by layer using GPT-4o.

\textbf{TaxoAdapt}~\cite{Kargupta_Zhang_Zhang_Zhang_Mitra_Han_2025} adaptively expands taxonomy width and depth according to the distribution of event types assigned to each node. Following the original setting, we use predefined domains and event-type inventories extracted from the corpus. GPT-4o first generates pseudo taxonomy labels, after which event types are assigned through classification-based matching. Width or depth expansion is then dynamically triggered according to node coverage.

\textbf{Human-Crafted Taxonomy.}
We include the CHED taxonomy~\cite{congcong-etal-2023-ched} as a human-crafted reference taxonomy. Curated from the \textit{Twenty-Four Histories} for historical event extraction, it provides a coarse-grained manually designed event hierarchy.

\section{Metrics}
\label{metrics}

\textbf{Path Granularity.} We use GPT-4o to score each parent–child pair, assigning 1 if the child is more specific than the parent and 0 otherwise. Finally, we compute the proportion of pairs that satisfy this condition.

\textbf{CSC}. CSC is a structure-level metric that quantifies the alignment between the taxonomy’s hierarchical structure and the semantic similarity of concepts. 
Specifically, it measures the extent to which concept pairs that are farther apart in the taxonomy are also semantically less similar, while closer nodes exhibit higher semantic similarity.

\begin{equation}
W_{n_a,n_b} = \frac{2 \cdot \text{lca}(p(n_a), p(n_b))}{|p(n_a)| + |p(n_b)|}
\end{equation}
where $p(n_a)$ denote the path from the root concept to a target concept $n_a$. Furthermore, let $\text{lca}(p(n_a),p(n_b))$ denotes the depth of the least common ancestor of the paths $p(n_a)$ and $p(n_b)$.
\begin{equation}
\text{CSC} := \tau \left( W_{n_a n_b}, S_{n_a n_b} \right)
\end{equation}
where $\tau$ denotes Kendall rank correlation, and $S_{n_a n_b}$ represents the semantic similarity between concept representation. We encode each node using \texttt{text-embedding-3-small} and compute cosine similarity to obtain $S$.

\textbf{Coverage Rate.} This metric evaluates how well the generated taxonomy can organize knowledge from the \textit{Twenty-Four Histories}.
We sample two chapters from each book (excluding those used during experiments), and use Qwen3, DeepSeek-V3, and GPT-4o to extract event samples (text + event type). After deduplication, 
we compute the cosine similarity between each event type and every node in the taxonomy. If the similarity is above 0.6, we consider that event type covered. The final coverage rate is the proportion of event types that are covered. All nodes are encoded using \texttt{text-embedding-3-small}.

\textbf{Node Recall.} This metric measures how many nodes in the human taxonomy ($\mathcal{T}$) are recalled by the generated taxonomy ($\mathcal{C}$). We encode each node in both taxonomies using \texttt{text-embedding-3-small} and compute cosine similarity between node pairs. If the similarity exceeds 0.6, we consider the human node recalled. 

\textbf{Novelty.} Novelty quantifies the proportion of completely new classes or categories in a taxonomy that are not present in any previous taxonomies. For a generated taxonomy ($\mathcal{C}$) and a human taxonomy ($\mathcal{T}$), it is defined as:

\begin{equation}
\small
\text{Novelty}(\mathcal{C}, \mathcal{T}) = \frac{\sum_{c \in \mathcal{C}} \text{new}(\mathcal{C}, \mathcal{T}, c)}{|\mathcal{C}|} \in [0,1]
\end{equation}

Here, ($\text{new}(\mathcal{C},\mathcal{T},c) = 1$) if class ($c$) is entirely new compared to taxonomy ($\mathcal{T}$), and 0 otherwise. A higher value indicates that the taxonomy introduces more novel information.

To determine whether a class $c$ is new relative to the human taxonomy, we use the same cosine-similarity procedure as in Node Recall. If a node in the generated taxonomy cannot find any human node with similarity greater than 0.6, we consider it a novel node.








\textbf{Significance.} We measure the significance of a taxonomy by its ability to provide fine-grained categorization of events.
To compare a generated taxonomy ( $\mathcal{C}$ ) with a human-curated taxonomy ( $\mathcal{T}$ ), we adopt classification delta, defined as:

\begin{equation}
\text{class.\ delta}(\mathcal{C}, \mathcal{T}, R) = \frac{|\sim_\mathcal{C}| - |\sim_\mathcal{T}|}{|R|}
\end{equation}

where ( $R$ ) is the set of event types extracted by LLMs during the Coverage Rate evaluation and successfully matched by both taxonomies.

Here,
 ($ |\sim_\mathcal{C}| $) denotes the number of distinct taxonomy nodes in ( $\mathcal{C}$ ) that events in ( $\mathcal{R}$ ) are assigned to,
( $|\sim_\mathcal{T}|$ ) denotes the corresponding number of distinct nodes in the human taxonomy ( $\mathcal{T} $).

A positive classification delta indicates that events in ( $R$ ) are distributed across more nodes in ( $\mathcal{C}$ ) than in ( $\mathcal{T}$ ), implying that ( $\mathcal{C}$ ) provides a more fine-grained categorization.
A delta of zero indicates that ( $\mathcal{C}$ ) matches the granularity of the human taxonomy, while a negative value suggests that ( $\mathcal{C}$ ) is coarser and assigns events to fewer categories.

\section{Structural Statistics of Taxonomies}
\label{taxonomy_statis}
The statistics of the taxonomies generated by different method are reported in Table~\ref{tab:struct_stats}.We report the maximum depth, average depth, and the average number of branches per parent node. We find that taxonomies generated by LLMs in a single run, namely Qwen3, DeepSeek-V3, and GPT-4o, tend to exhibit relatively simple structures, characterized by shallower depths. In contrast, the human-crafted taxonomy and the multi-agent method produce structures that are generally deeper and more hierarchical.

\begin{table*}[h]
  \centering
  \small
\caption{Structural statistics of taxonomies generated by different methods.}
  \label{tab:struct_stats}
  \setlength{\tabcolsep}{0.08mm}
  \begin{tabular}{c|c|cc|ccc}
    \hline
    \Xhline{1.2pt}
    Models & Strategies &  Nodes &  Leaves & Max. Dep. & Avg. Dep. & Branch \\
    \hline
    
    Qwen3 & \multirow{3}{*}{Domain} 
    & 112 & 72 & 3 & 3 & 3 \\
    DeepSeek-V3 & 
    & 103 & 63 & 3 & 3 & 2.72 \\
    GPT-4o & 
    & 112 & 72 & 3 & 3 & 3 \\
    \hline
    
    Qwen3 & \multirow{3}{*}{Corpus} 
    & 113 & 72 & 3 & 3 & 3 \\
    DeepSeek-V3 & 
    & 74 & 53 & 3 & 3 & 4 \\
    GPT-4o & 
    & 60 & 40 & 4 & 4 & 3.05 \\
    \hline
    
    Qwen3 & \multirow{3}{*}{Corpus + Domain} 
    & 96 & 56 & 3 & 3 & 2.5 \\
    DeepSeek-V3 & 
    & 214 & 180 & 3 & 3 & 7.62 \\
    GPT-4o & 
    & 96 & 58 & 3 & 3 & 2.67 \\
    \hline

    HiExpan & Corpus + Domain
    & 208 & 157 & 6 & 4.52 & 4.65 \\
    \hline
    
    HiSetExpan & Corpus + Domain
    & 118 & 86 & 4 & 3.81 & 4.58 \\
    \hline
    
    \multirow{2}{*}{CoL} 
    & Domain
    & 540 & 440 & 6 & 3.74 & 5.75 \\
    \cline{2-7}
    & Corpus + Domain
    & 160 & 108 & 6 & 3.67 & 3.27 \\
    \hline
    
    TaxoAdapt & Corpus + Domain
    & 1109 & 860 & 6 & 4.93 & 4.59 \\
    \hline
    
    CHED & Human
    & 100 & 67 & 5 & 2.73 & 3.52 \\
    \hline
    
    CHisAgent & Corpus + Domain
    & 483 & 283 & 6 & 3.93 & 2.43 \\
    
    \hline
    \Xhline{1.2pt}
  \end{tabular}

  \vspace{-5mm}
\end{table*}

\section{Additional Domain Analysis}
\label{appendix_domain_analysis}

\begin{figure}[ht]
    \centering
        \subfigure[Path Granularity.] {
    \label{f2a}       
\includegraphics[width=0.45\columnwidth]{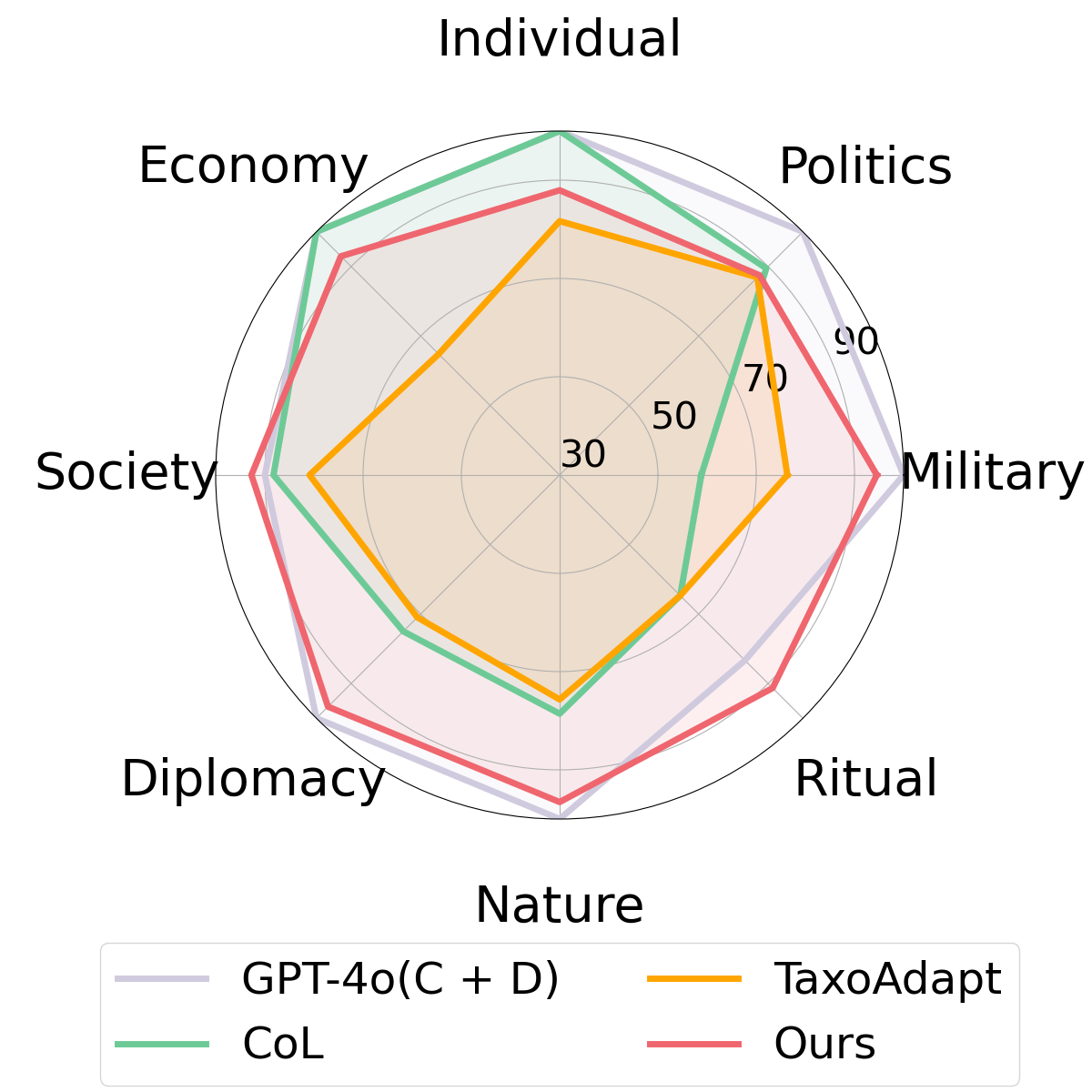}  
    } 
    \subfigure[Node Recall.] {
     \label{f2c}     
\includegraphics[width=0.45\columnwidth]{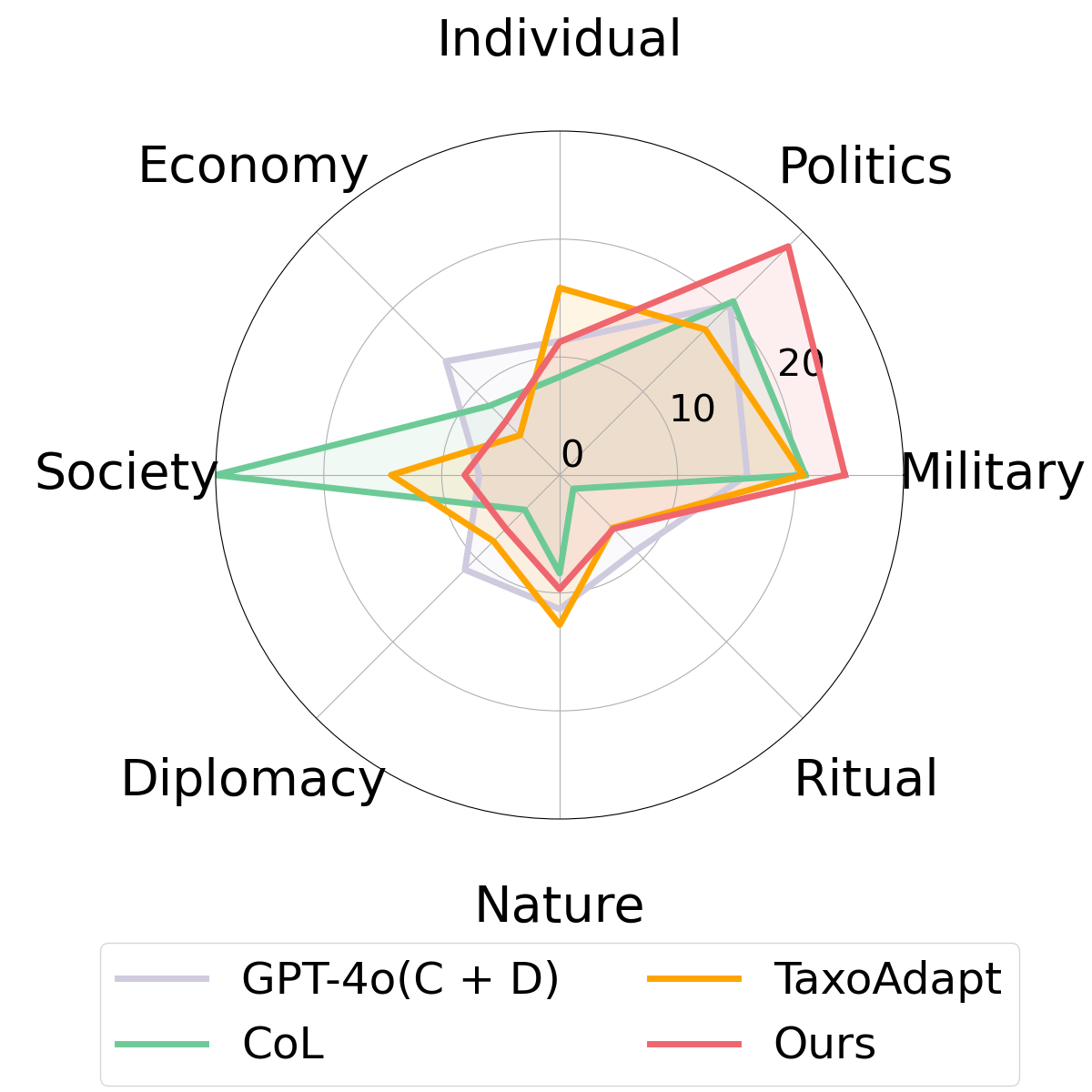}  
    } 
    \vspace{-3mm}
    \caption{Radar chart of evaluation metrics of different methods across various domains. GPT-4o (C + D) denotes GPT-4o (Corpus + Domain), and ``Economy'' represents ``Economy-Livelihood''.}
    \label{f7}
     \vspace{-3mm}
\end{figure}

Figure~\ref{f2a} presents parent--child relationship rationality. Our method and GPT-4o consistently achieve relatively high scores across all domains, whereas CoL and TaxoAdapt show larger performance fluctuations, particularly in Ritual and Diplomacy.
Figure~\ref{f2c} illustrates the distribution of recalled nodes. Different methods exhibit distinct preferences: our method mainly recalls nodes from Politics and Military; TaxoAdapt focuses more on Military, Society, Nature, and Individual; and CoL concentrates on Military, Society, and Diplomacy.

\section{Human Evaluation}
\label{guideline}
\textbf{Node Clarity.}
For each taxonomy, we randomly sample 50 nodes and ask evaluators to judge whether the node’s semantics are clear and whether the node can appropriately represent an event type in ancient Chinese cultural system. Scores range from 0 to 5.

\textbf{Hierarchical Coherence.}
For each taxonomy, we randomly sample 50 nodes together with their full paths, and evaluators assess whether the hierarchical structure from the root to the leaf is logically coherent. Scores range from 0 to 5.

\textbf{In-domain Orthogonality.}
For each taxonomy, we randomly sample 50 nodes and retrieve their top 5 most similar nodes within the same domain. Evaluators then assess whether the semantics of the sampled node overlap with those similar nodes. If no semantic overlap exists, the node receives a score of 5; if there is overlap with one node, the score is 4; and so on.

\textbf{Cross-domain Orthogonality.}
The procedure is similar to in-domain orthogonality, except that the top 5 similar nodes are retrieved from other domains. Evaluators judge whether cross-domain semantic overlap exists.

\textbf{Completeness.}
Evaluators assess whether a taxonomy sufficiently covers the conceptual space of Chinese historical events, considering its overall breadth, depth, and coverage of important event types and conceptual categories. The evaluation is conducted at the taxonomy level to better reflect the practical usefulness and comprehensiveness of the constructed taxonomy. Scores range from 0 to 5.

\begin{figure*}
    \centering
    \small
    \includegraphics[width=0.9\linewidth]{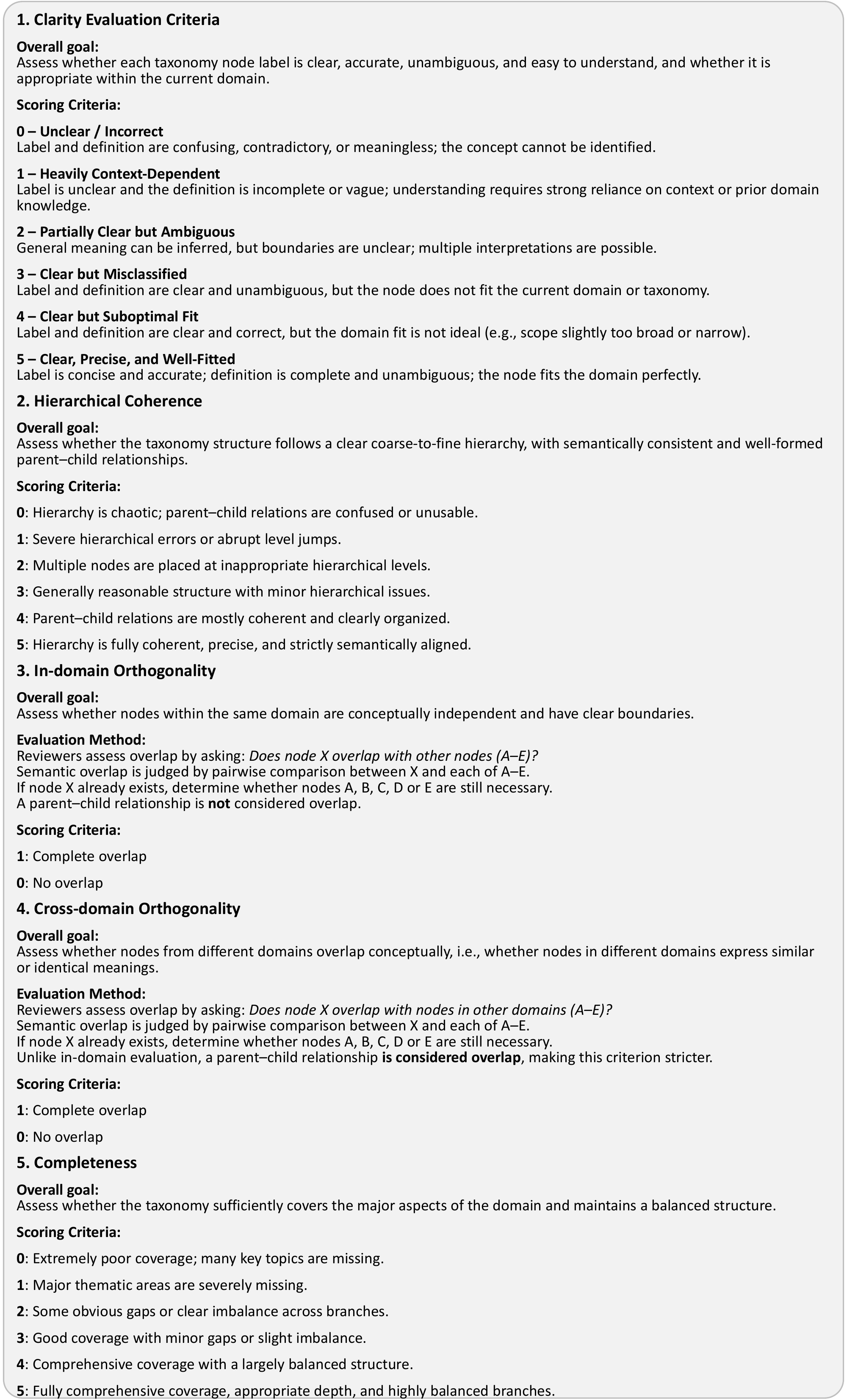}
    \caption{Guidelines for Human Evaluation.}
    \label{guidelines}
\end{figure*}

All human evaluation guidelines are presented in Figure~\ref{guidelines}.

\section{Error Analysis}
\label{appendix_error_analysis}

We conduct a quantitative error analysis based on the human evaluation results (Section~\ref{human_eval}). Specifically, we analyze the proportion of nodes that received low ratings across different evaluation dimensions:

\begin{itemize}

    \item \textbf{Node Clarity}: 1/50 sampled nodes received a score lower than 4.

    \item \textbf{Hierarchical Coherence}: 16/50 sampled nodes received a score lower than 4.

    \item \textbf{In-domain Orthogonality}: 10/50 sampled nodes were judged to exhibit semantic redundancy (lower than 5) with other nodes in the same domain.

    \item \textbf{Cross-domain Orthogonality}: 4/50 sampled nodes exhibited semantic overlap (lower than 5) across domains.

\end{itemize}

Completeness evaluates the overall coverage of the taxonomy rather than individual nodes and therefore is not suitable for node-level proportion analysis.

These results indicate that while our method maintains strong semantic clarity, the primary limitations lie in hierarchical organization and intra-domain redundancy.

To further illustrate these error types, we provide representative examples below:

\subsection{Hierarchical Coherence Errors}

The main limitation is over-segmentation and redundant intermediate layers.

In some cases, the model introduces overly fine-grained or semantically redundant intermediate nodes, resulting in weak hierarchical justification. For example:

\begin{itemize}

    \item 政治 (Politics) $\rightarrow$ 人事任免 (Personnel Appointment) $\rightarrow$ 任用 (Appointment) $\rightarrow$ 任命 (Nomination) $\rightarrow$ 職務調整 (Position Adjustment)

    \item 政治 (Politics) $\rightarrow$ 政務與決策 (Governance and Decision-making) $\rightarrow$ 決策裁定 (Decision Ruling) $\rightarrow$ 決策 (Decision)

    \item 個人 (Individual) $\rightarrow$ 行為活動 (Behavioral Activities) $\rightarrow$ 辯解 (Justification) $\rightarrow$ 解釋 (Explanation)

    \item 個人 (Individual) $\rightarrow$ 行為活動 (Behavioral Activities) $\rightarrow$ 允諾 (Promise) $\rightarrow$ 同意 (Agreement)

\end{itemize}
In these examples, parent and child nodes are nearly synonymous, and the additional layers do not introduce meaningful semantic distinctions. This suggests that the model occasionally over-expands the hierarchy without sufficient semantic differentiation.

\subsection{In-domain Orthogonality Errors}

To further analyze semantic redundancy, we examined sampled nodes flagged during in-domain orthogonality evaluation. Among the 10 sampled cases with semantic overlap, only one node received a low final score (2/5), while most cases received 4/5, indicating mild rather than severe redundancy.


Several overlapping cases occur within ritual-related categories, such as:

\begin{itemize}

    \item 祭祀与皇室活動(Ritual and Imperial Activities)

    \item 祭祀活動實施 (Ritual Implementation)

    \item 定期祭祀 (Regular Ritual)

    \item 祈禳祭祀 (Supplicatory Ritual)

    \item 祭祀供品与物品 (Ritual Offerings)

    \item 儀式執行 (Ceremonial Execution)

\end{itemize}

These nodes differ along multiple dimensions (function, frequency, material components, and execution process). However, the taxonomy occasionally mixes these classification criteria, leading to \textit{partial semantic overlap}. This suggests that the model does not always consistently align subtype expansion along a single ontological dimension.

\subsection{Cross-domain Orthogonality Errors}

We further examine cross-domain semantic overlap cases. Among the sampled nodes, only one case received a low final score (2/5), while most received 4/5, indicating that severe cross-domain redundancy is rare.

We identify a recurring pattern: institutional abstraction across domains.

Certain nodes involve institutional or regulatory structures that naturally exist across multiple domains. For example:

\begin{itemize}

    \item 外交 (Diplomacy) $\rightarrow$ 報告制度 (Reporting Policies)

    \item 政治 (Politics) $\rightarrow$ 制度与政策 (Institutional Reform / Policies)

\end{itemize}
Although categorized under different domains, these nodes share a higher-level abstraction (“institutional mechanism”), which leads to partial semantic proximity. This reflects the inherent cross-domain nature of institutional governance structures.

\section{Case Study}
 In this section, we add detailed running examples to illustrate the role of each stage, Inducer, Expander, and Enricher, in a step-by-step manner, as described below.

\begin{figure*}
    \centering
    \includegraphics[width=14cm]{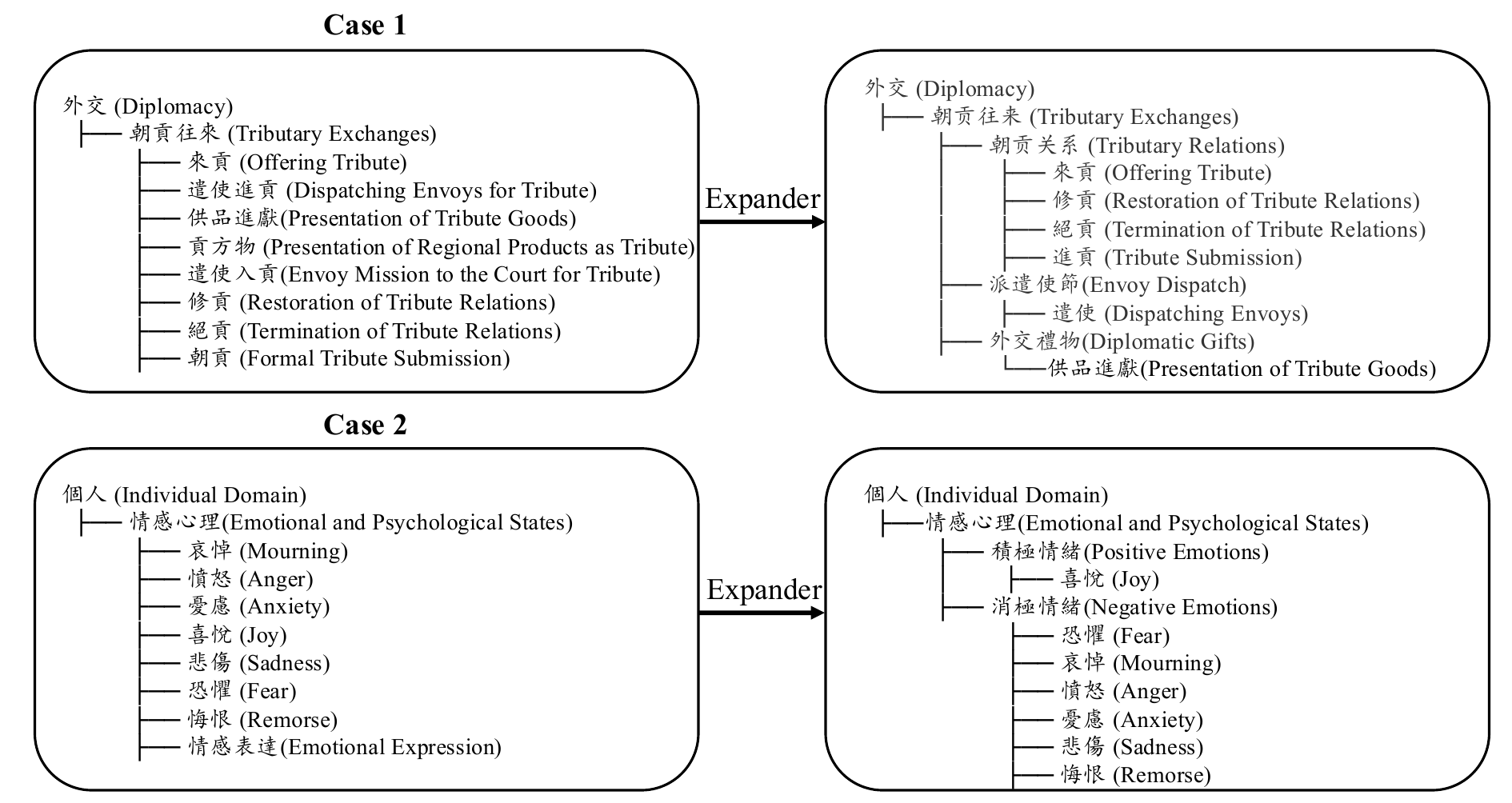}
    \caption{Expander examples in Diplomacy and Individual domains.}
    \label{case_expander}
\end{figure*}

 As shown in Figure~\ref{case_expander}, induced taxonomy (after Inducer) are at left. In the induced taxonomy, all tribute-related actions were placed at the same level. This resulted in a mixture of: relational states (e.g., Restoration of Tribute Relations, Termination of Tribute Relations), operational actions (e.g., Envoy Mission to the Court for Tribute), and ritual gift exchanges (e.g., Presentation of Tribute Goods). After applying the Expander: The Expander introduces three intermediate categories: 朝貢關係 (Tributary Relations) (relational dimension), 使節派遣 (Envoy Dispatch) (operational dimension), 外交禮物(Diplomatic Gifts) (ritual/material exchange dimension).
This restructuring improves sibling-level semantic coherence and clarifies the institutional logic of tribute diplomacy. The example demonstrates that the Expander does not merely add nodes, but reorganizes heterogeneous siblings into conceptually consistent substructures.

We further illustrate structural refinement in the ``Individual – Emotional'' domain in Figure~\ref{case_expander}. Here, emotional states of different polarity were placed at the same level, resulting in a flat and heterogeneous structure.

After applying the Expander, the Expander introduces a polarity-based abstraction (Positive Emotions vs. Negative Emotions), enforcing conceptual consistency between parent and child nodes. This reduces semantic heterogeneity and enhances structural clarity.

\begin{figure}
    \centering
    \includegraphics[width=7cm]{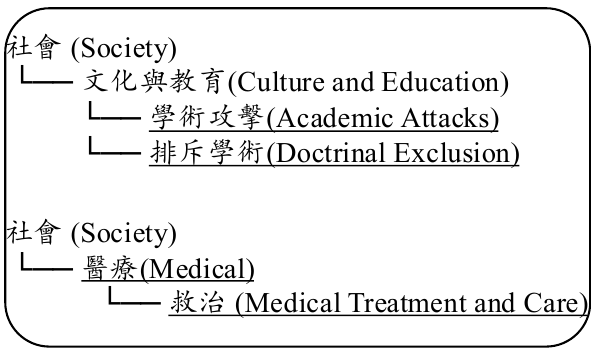}
    \caption{Enricher examples in Society domains.}
    \label{case_enricher}
\end{figure}

To illustrate the Enricher stage, we provide examples where important event types were absent from the induced taxonomy and were subsequently introduced through external knowledge and corpus signals. As shown in Figure~\ref{case_enricher}, 
For the case 1, the initial taxonomy lacked categories capturing intellectual conflicts. However, historical records frequently public denunciations, doctrinal exclusion, suppression of teaching, factional academic disputes.
The Enricher identifies and incorporates these discourse-level events, adding 學術攻擊 (Academic Attacks) and 排斥學說(Doctrinal Exclusion) under the cultural domain. These nodes capture dynamics of intellectual power and doctrinal boundary-making that were previously unrepresented.

In the second case, the initial taxonomy did not include any medical-related events. However, historical documents frequently describe actions such as diagnosis, prescribing medicine, and treatment outcomes. The Enricher introduces ``醫療(Medical)-救治 (Medical Treatment and Care)'' to improve coverage of everyday social practices.

This example demonstrates that the Enricher performs semantic coverage expansion rather than structural reorganization.

\section{Taxonomy Example}
\label{example}
\begin{figure*}[h]
\centering
 \subfigure[Diplomacy.] {
     \label{f5a}     
\includegraphics[width=0.9\columnwidth]{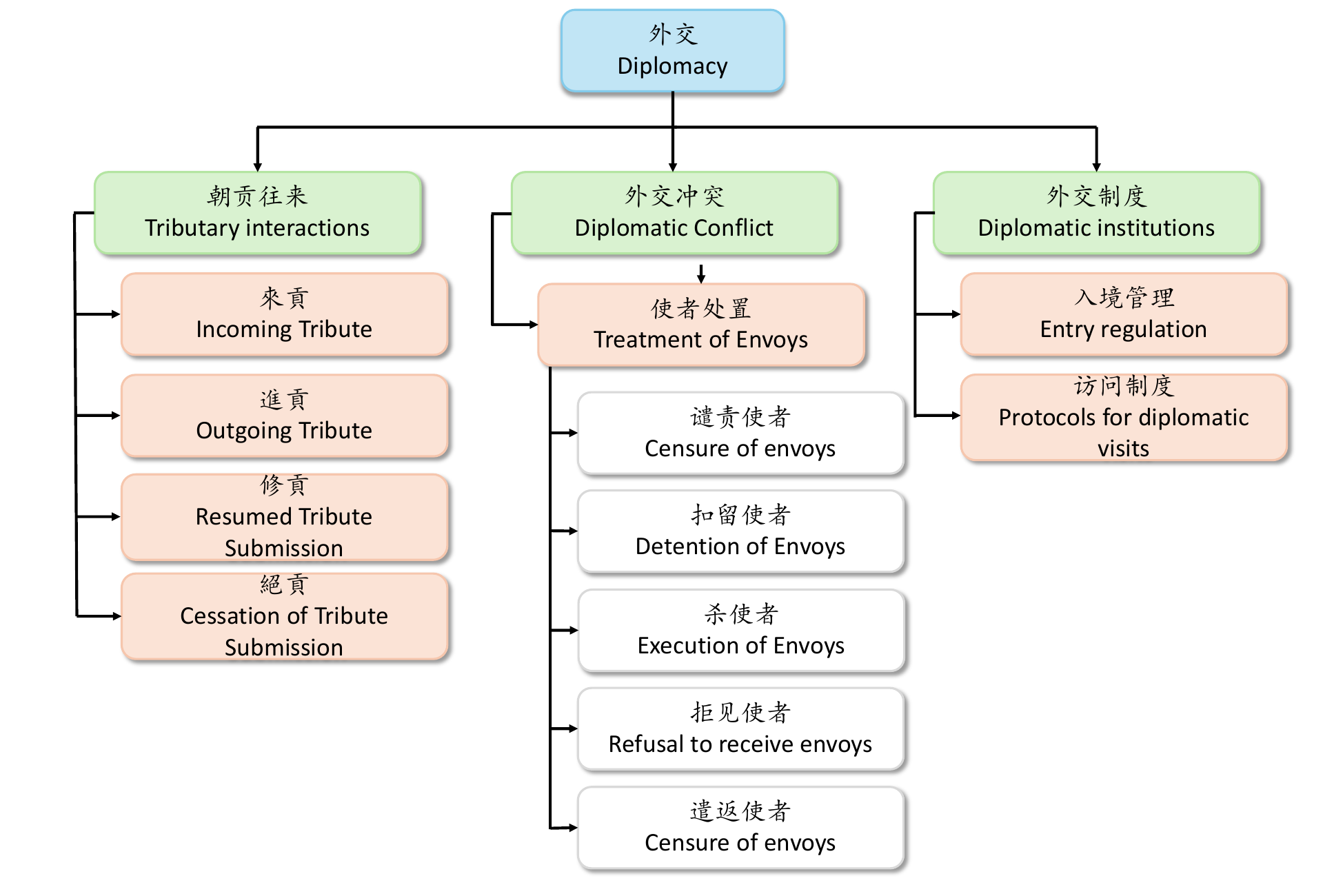}  
    }   
     \subfigure[Society.] {
     \label{f5a}     
\includegraphics[width=0.9\columnwidth]{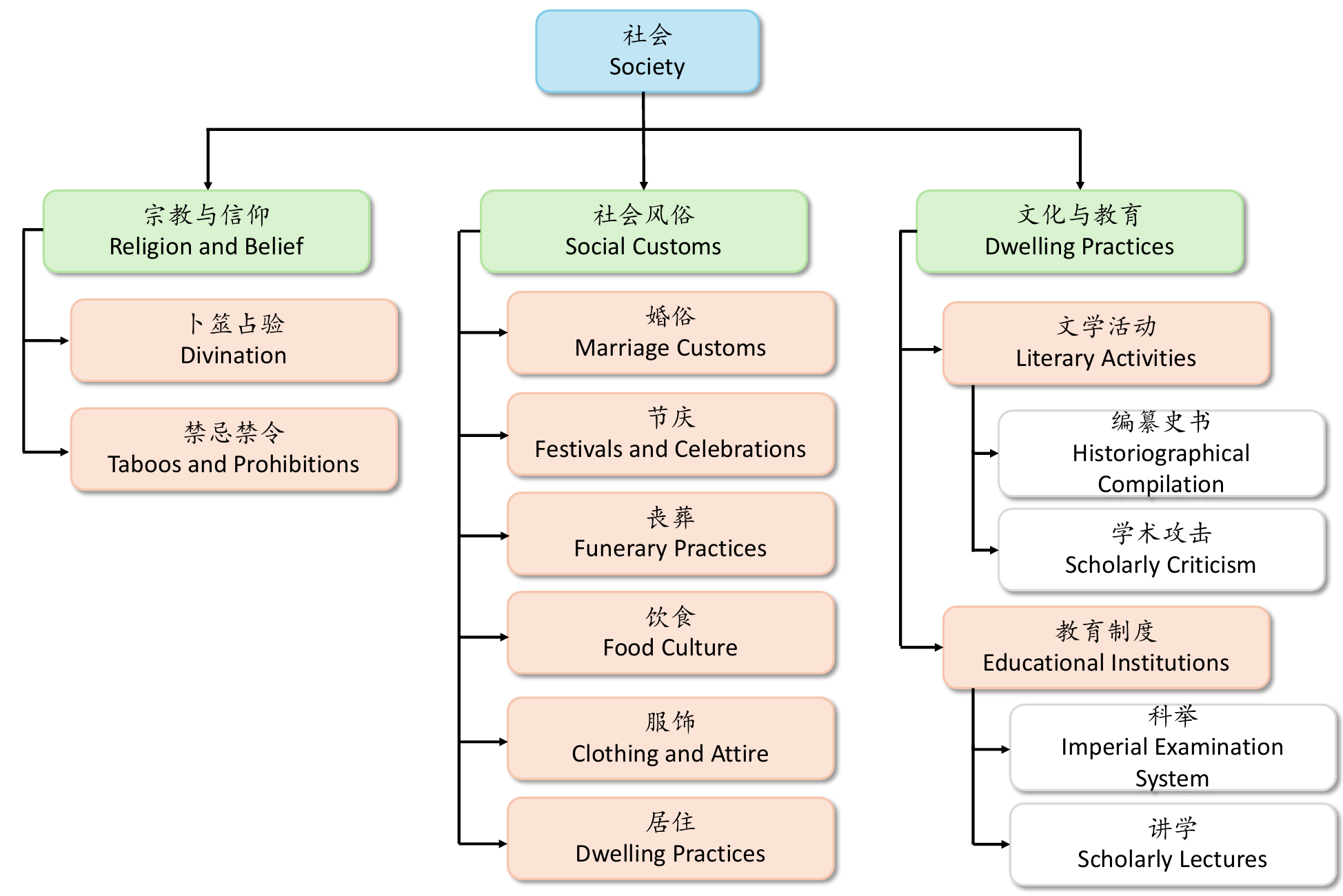}  
    }   
    \caption{Taxonomy (partial) Example for Diplomacy and Society Domain.}
    \label{example_prompt}
\end{figure*}

Figure~\ref{example_prompt} illustrates representative portions of the core taxonomy for the diplomacy and society domains.

\section{Sensitivity to Similarity Thresholds}
To examine the sensitivity of the proposed methods to the similarity threshold, we evaluate all methods under five threshold settings, ranging from 0.5 to 0.7. The results are reported in Table~\ref{tab:threshold}. Although the absolute values of Coverage, Node Recall, and Novelty vary with different thresholds, the relative performance trends across methods remain largely consistent. In particular, CHisAgent and TaxoAdapt consistently outperform CoL in terms of Coverage and Node Recall across different threshold settings, while achieving a favorable balance with Novelty. These results demonstrate that the comparative performance of the methods is robust to the choice of similarity threshold. Based on this analysis, we retain 0.6 as the default threshold for our experiments.
\begin{table}[t]
\centering
\small
\caption{Performance comparison under different similarity thresholds.}
\label{tab:threshold}
\setlength{\tabcolsep}{0.2mm}
\begin{tabular}{lcccc}
\toprule
\textbf{Methods} & \textbf{Threshold} & \textbf{Coverage} & \textbf{Node Recall} & \textbf{Novelty} \\
\midrule
\multirow{5}{*}{\textbf{CoL}}
& 0.50 & 79.48 & 67.78 & 36.81 \\
& 0.55 & 58.12 & 55.11 & 43.75 \\
& 0.60 & 43.72 & 46.67 & 50.69 \\
& 0.65 & 34.84 & 28.89 & 70.83 \\
& 0.70 & 19.10 & 26.67 & 81.94 \\
\midrule
\multirow{5}{*}{\textbf{TaxoAdapt}}
& 0.50 & 98.66 & 85.56 & 44.69 \\
& 0.55 & 80.15 & 81.11 & 61.04 \\
& 0.60 & 74.17 & 60.00 & 69.21 \\
& 0.65 & 65.58 & 57.78 & 85.10 \\
& 0.70 & 51.76 & 44.44 & 92.55 \\
\midrule
\multirow{5}{*}{\textbf{CHisAgent}}
& 0.50 & 97.24 & 91.11 & 55.63 \\
& 0.55 & 88.11 & 73.33 & 68.05 \\
& 0.60 & 75.13 & 68.89 & 77.70 \\
& 0.65 & 66.16 & 45.56 & 87.82 \\
& 0.70 & 40.03 & 36.67 & 91.03 \\
\bottomrule
\end{tabular}
\end{table}

\section{Human Evaluation of Event Classification}

To independently assess the reliability of the Classifier agent, we conducted a human evaluation of its domain assignments. We randomly sampled \textbf{500 events }classified by GPT-4o and asked two annotators with expertise in Chinese historical texts to assign each event to one or more of eight predefined domains: Military (军事), Politics (政治), Diplomacy (外交), Society (社会), Individual (个人), Economy and Livelihood (经济民生), Nature (自然), and Ritual (祭祀). Annotators were allowed to assign multiple domains when an event involved multiple aspects.

The two annotators achieved substantial agreement despite the inherent ambiguity of historical domain classification. The exact label-set agreement was 59.6\%, with a mean Jaccard similarity of 0.71 and a macro-averaged Cohen’s $\kappa$ of 0.63. The per-domain $\kappa$ values ranged from 0.30 for Society to 0.81 for Military, indicating reasonable consistency across most domains.

We constructed a human consensus gold standard using the following procedure. When both annotators agreed on the primary domain, that domain was adopted as the consensus label. When their label sets had exactly one overlapping domain, the overlapping domain was adopted. Cases without a clear consensus ($N=86$, 17.2\%) were marked as disputed and excluded from strict evaluation, resulting in 414 events with reliable consensus labels.

Compared with this human-annotated gold standard, the Classifier agent achieved a strict accuracy of 68.4\% (283/414) on the consensus subset and a weighted F1 of 0.71. We additionally report a lenient accuracy of 79.6\% (398/500), which considers a prediction correct when the predicted domain falls within the union of domains assigned by the two annotators. This metric accounts for legitimate cross-domain interpretations and shows that many predictions remain within the range of human-acceptable classifications.

Error analysis indicates that most classification errors occur at the boundary between Politics (政治) and Individual (个人). Examples include ``爱民'' (showing benevolence toward the people), ``赐毒酒'' (granting poisoned wine), and ``科举及第'' (passing the imperial civil service examination). Such cases reflect genuine ambiguity between state-level affairs and individual actions in historical narratives, rather than straightforward classification failures.

These results provide an independent human-verified assessment of the Classifier agent. GPT-4o is used as the operational classifier in our pipeline, while the human evaluation serves as an external assessment of its classification reliability.

\section{Prompts}
\begin{figure*}[h]
    \centering
    \includegraphics[width=1\linewidth]{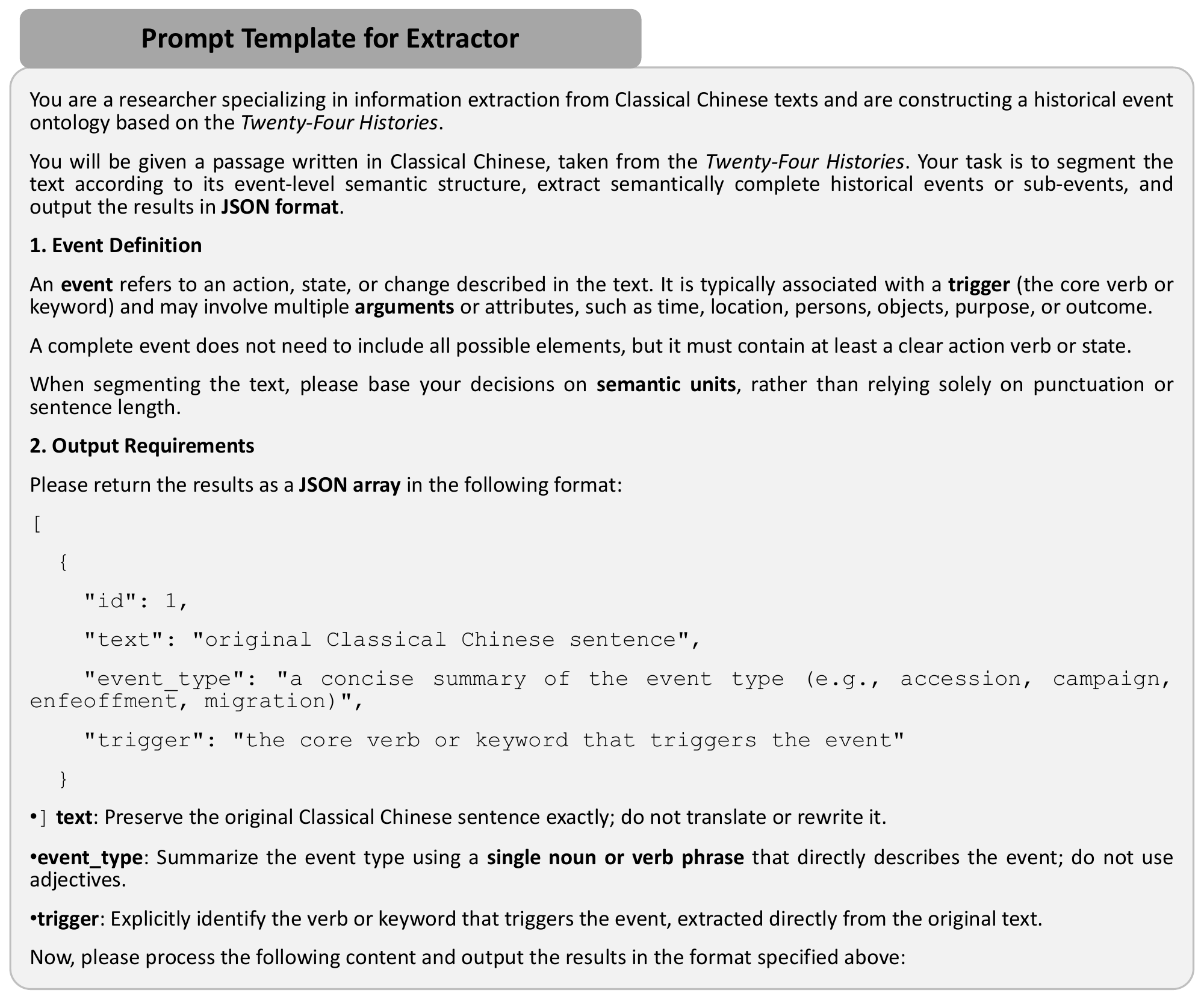}
    \caption{Prompt 1 – Extractor Prompt Example}
    \label{extractor_prompt}
\end{figure*}

\begin{figure*}[h]
    \centering
    \includegraphics[width=1\linewidth]{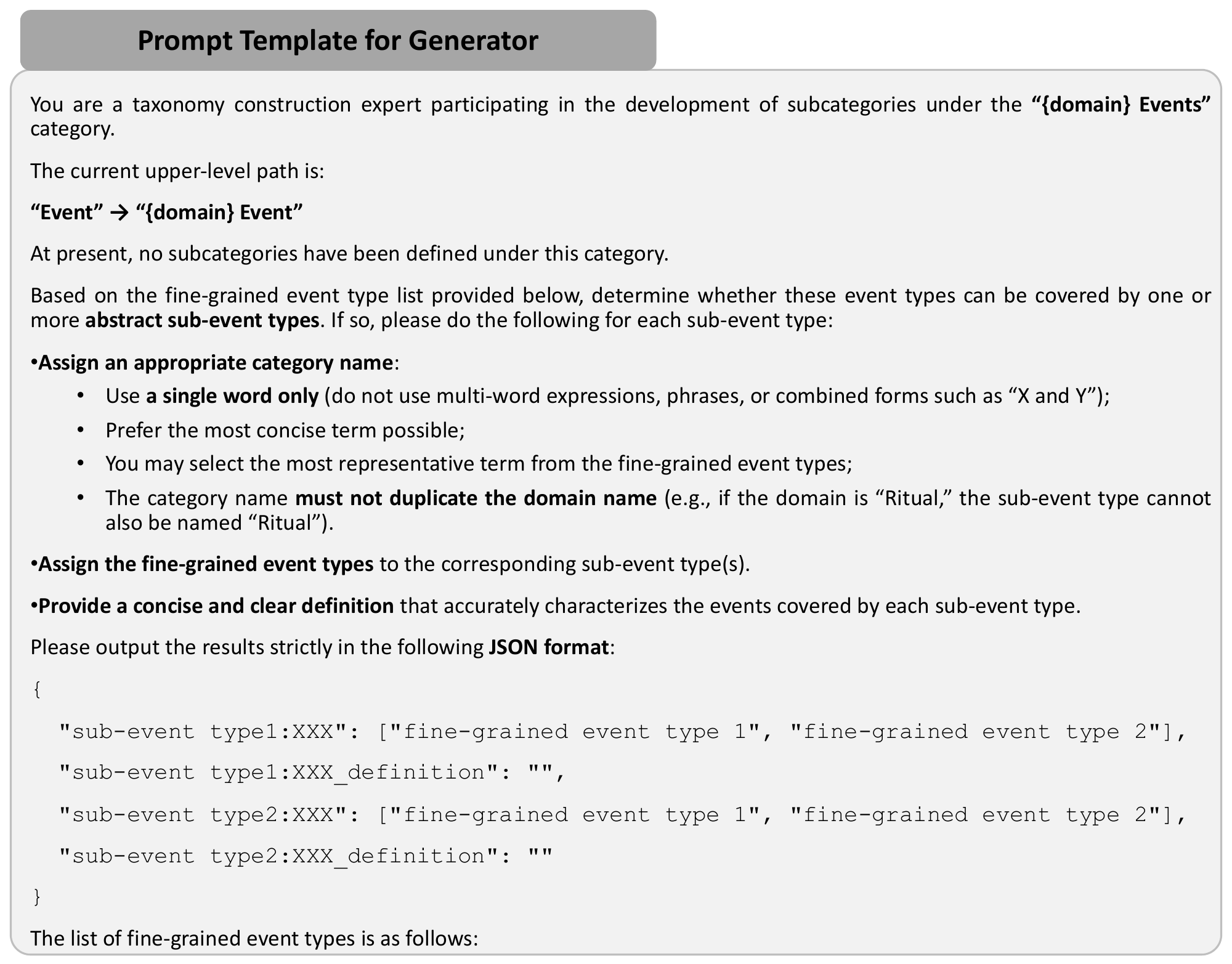}
    \caption{Prompt 3 – Generator Prompt Example}
    \label{generator_prompt}
\end{figure*}

\begin{figure*}[h]
    \centering
    \includegraphics[width=1\linewidth]{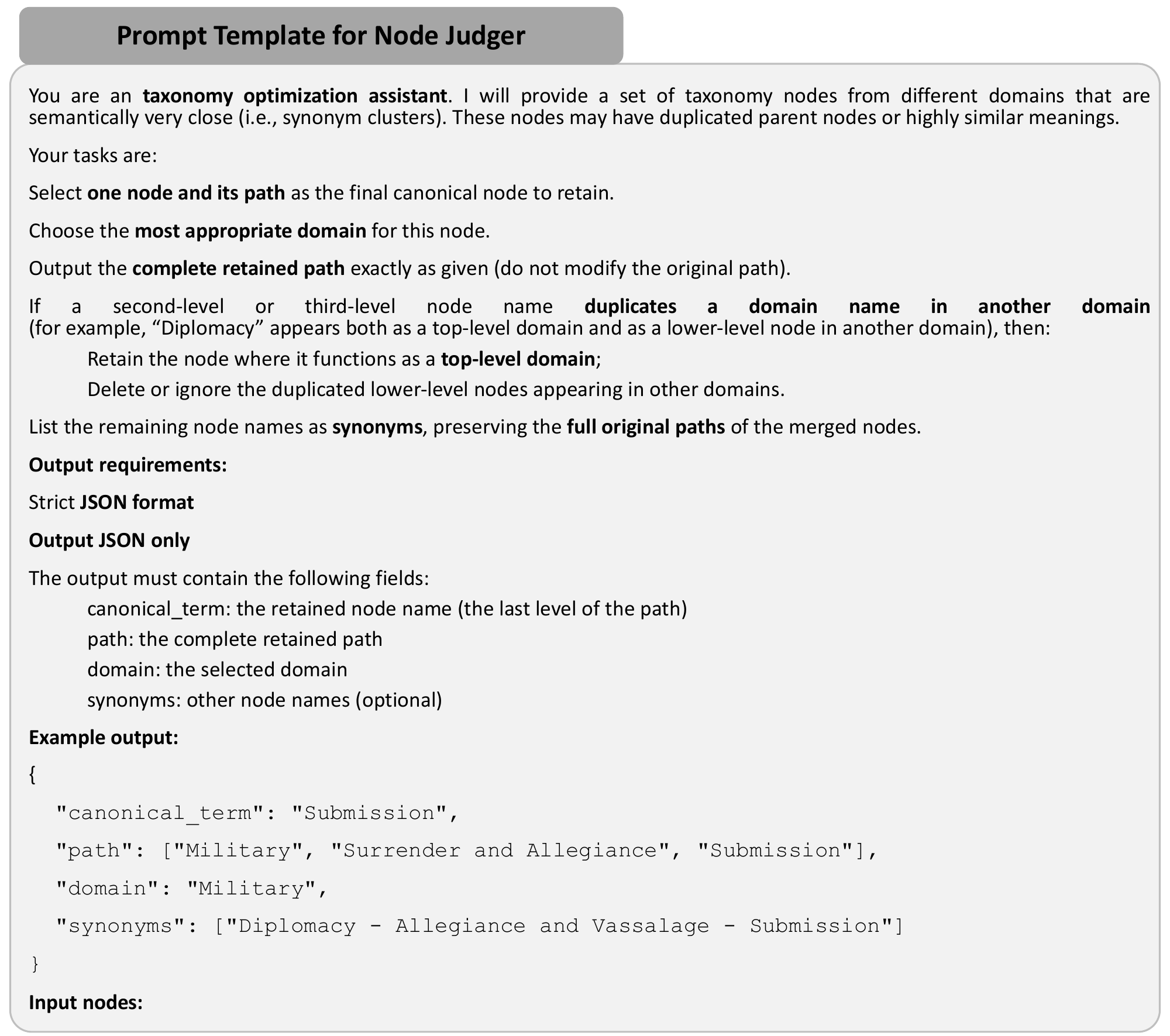}
    \caption{Prompt 5 – Node Judger Prompt Example}
    \label{remove_prompt}
\end{figure*}

\begin{figure*}[h]
    \centering
    \includegraphics[width=0.9\linewidth]{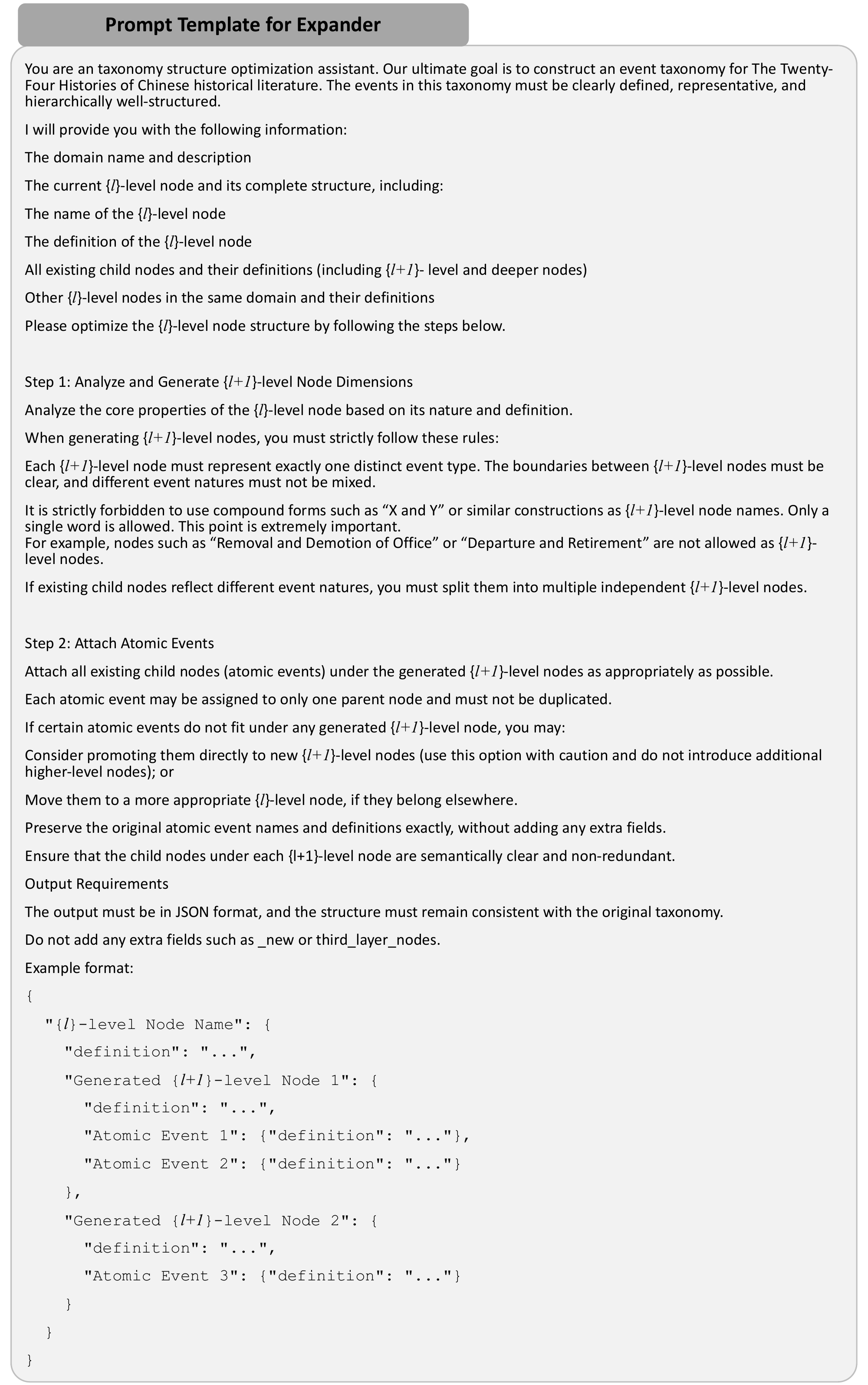}
    \caption{Prompt 6 – Expander Prompt Example}
    \label{expander_prompt}
\end{figure*}

\begin{figure*}[h]
    \centering
    \includegraphics[width=1\linewidth]{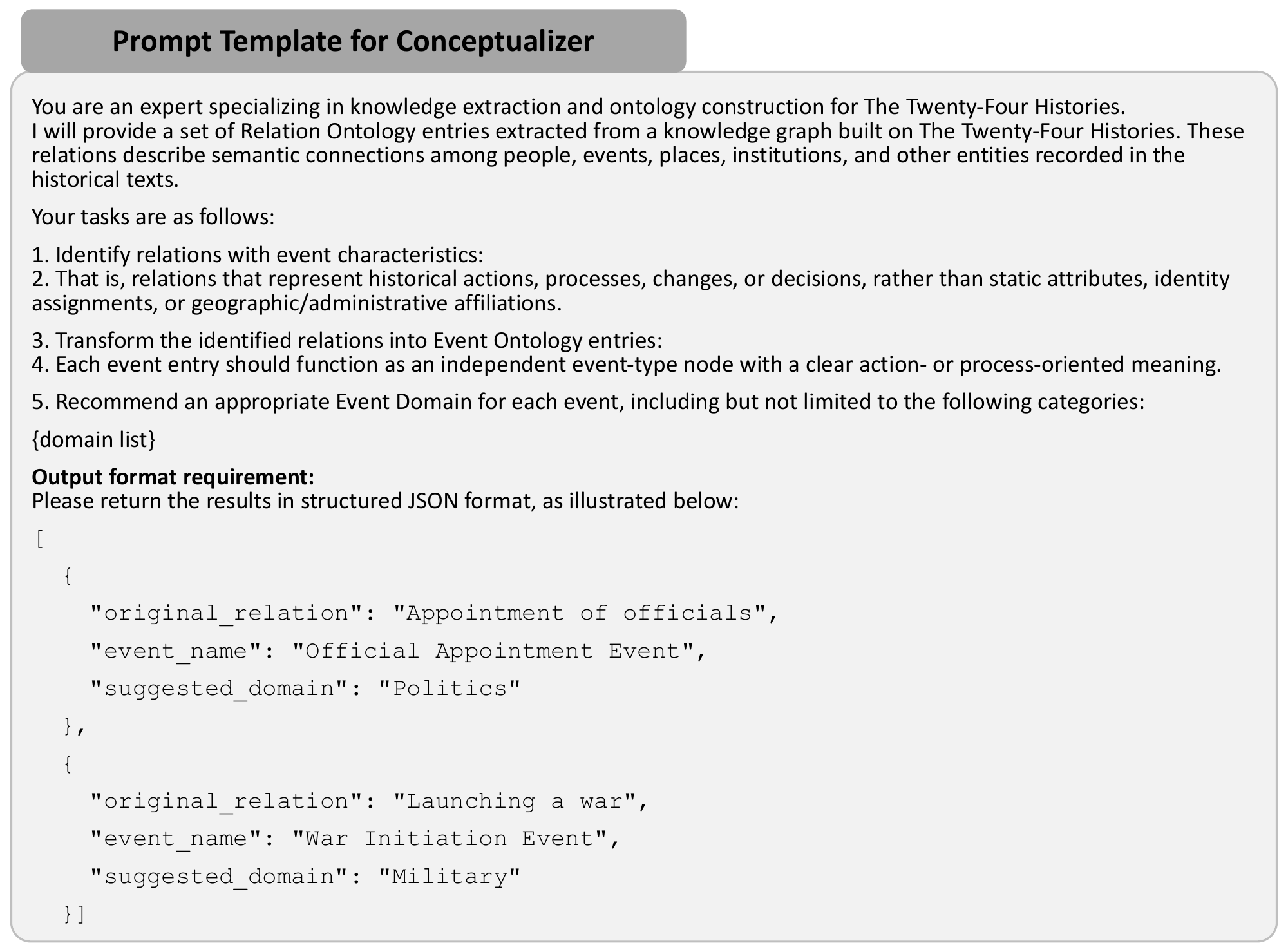}
    \caption{Prompt 7 – Conceptualizer Prompt Example}
    \label{conceptualizer_prompt}
\end{figure*}

\begin{figure*}[h]
    \centering
    \includegraphics[width=1\linewidth]{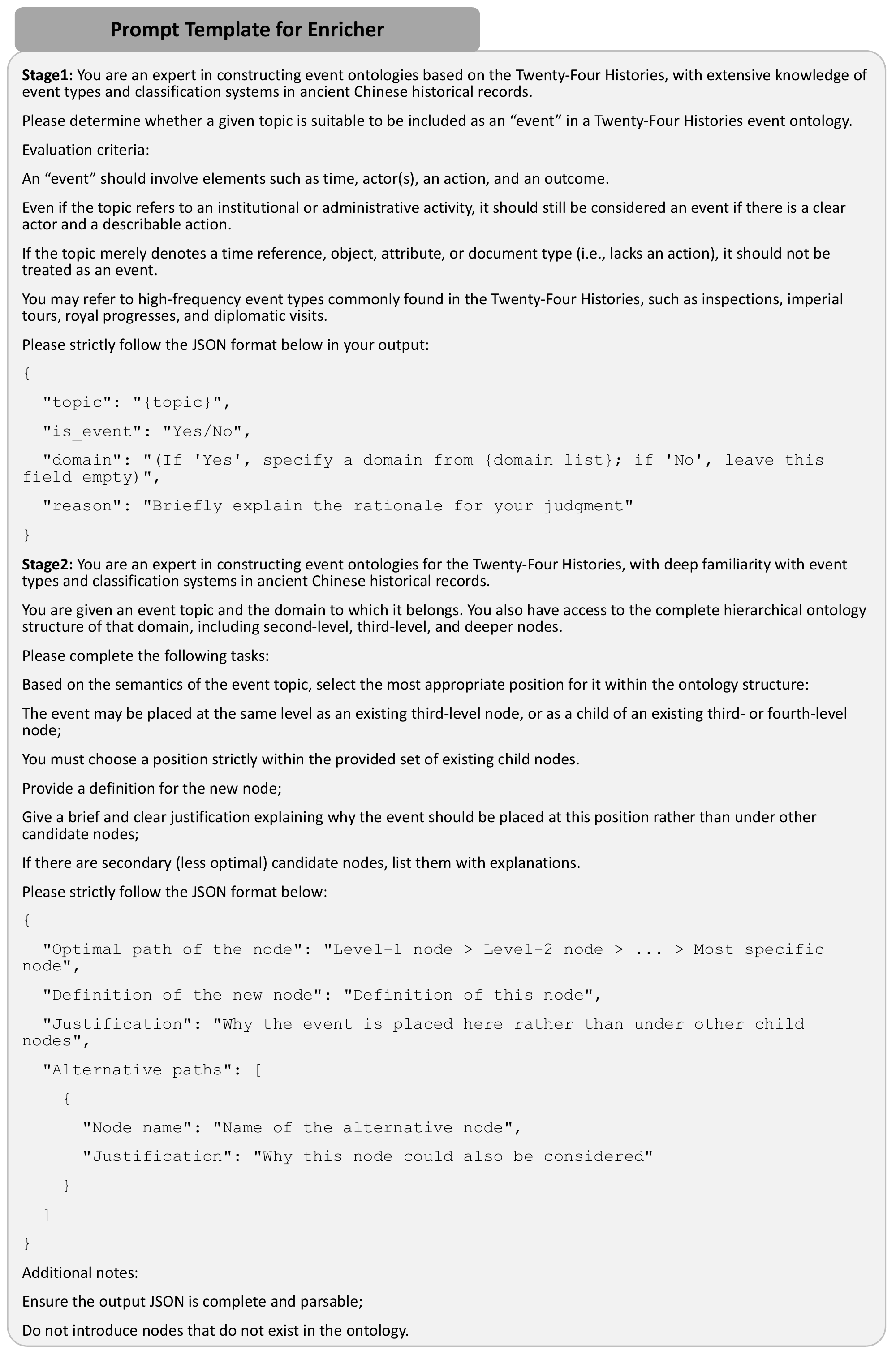}
    \caption{Prompt 8 – Enricher Prompt Example}
    \label{enricher_prompt}
\end{figure*}

All prompts used in the experiments are shown in Figures~\ref{extractor_prompt}–\ref{enricher_prompt}.
The original prompts are translated into English to facilitate understanding.
\begin{figure*}[h]
    \centering
    \includegraphics[width=1\linewidth]{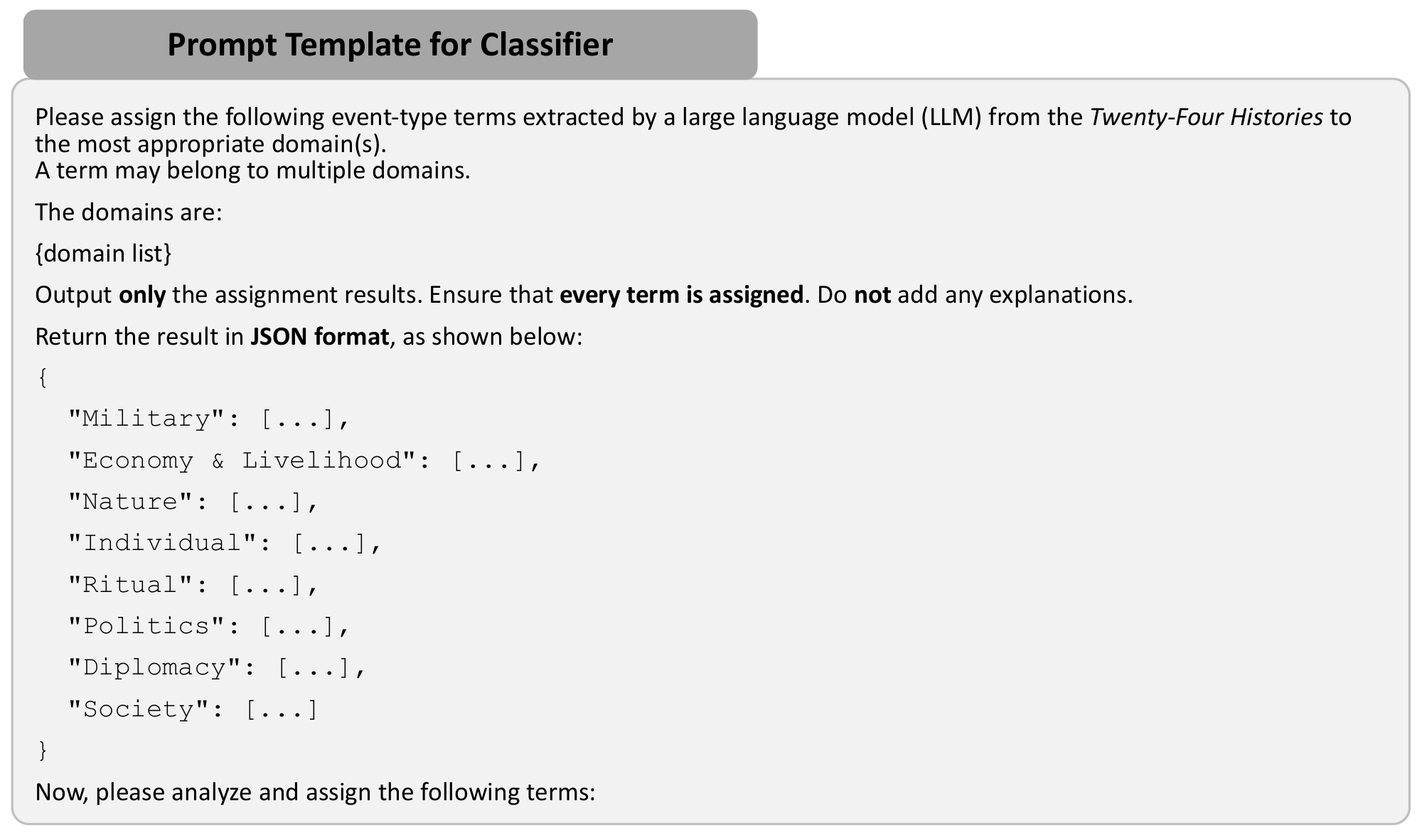}
    \caption{Prompt 2 – Classifier Prompt Example}
    \label{classifier_prompt}
\end{figure*}

\begin{figure*}[h]
    \centering
    \includegraphics[width=1\linewidth]{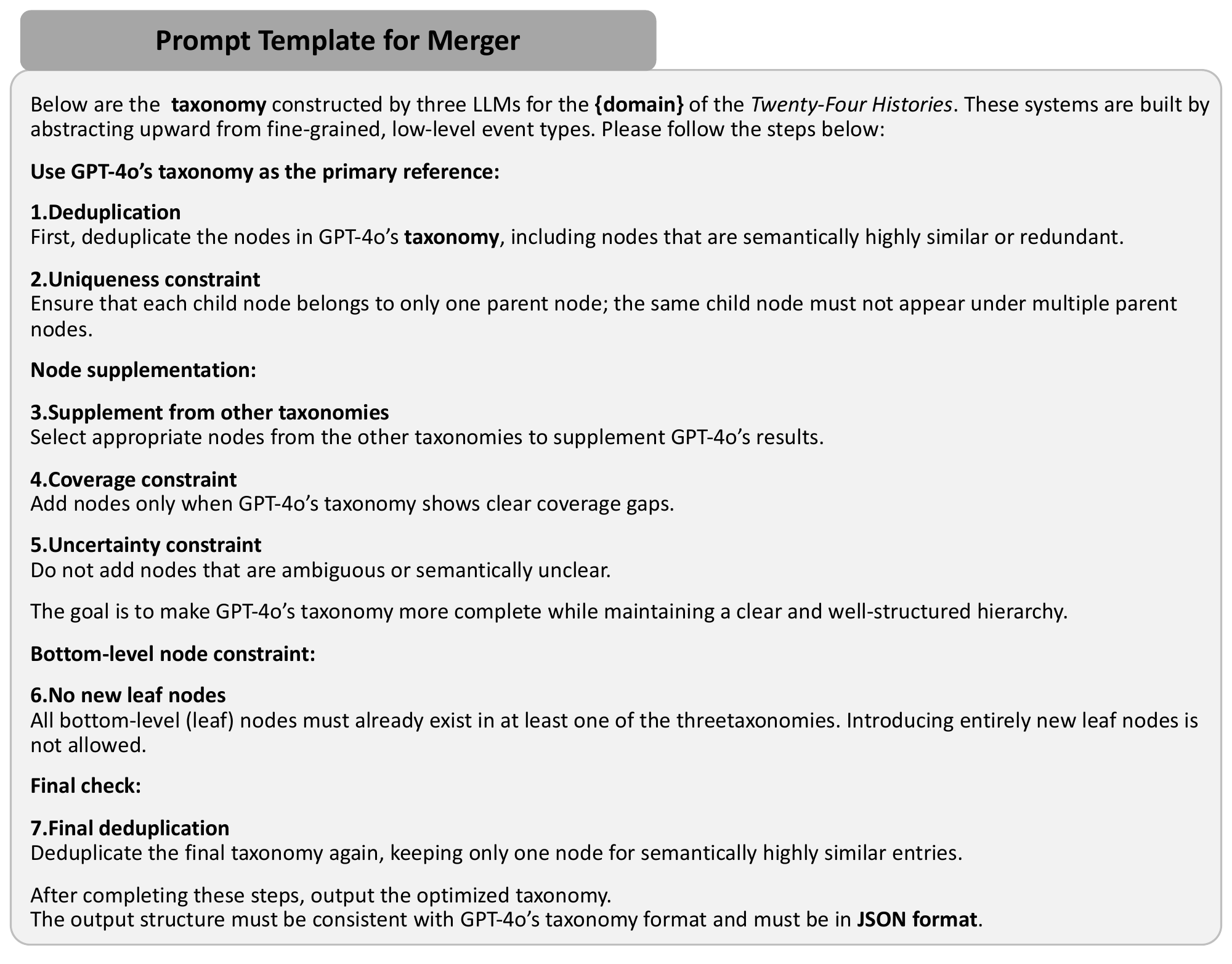}
    \caption{Prompt 4 – Merger Prompt Example}
    \label{classifier_prompt}
\end{figure*}

\end{CJK}
\end{document}